\documentclass[10pt,twocolumn,letterpaper]{article}

\usepackage[pagenumbers]{cvpr} 

\RequirePackage{algorithm}
\RequirePackage{algorithmic}

\usepackage{bm}
\usepackage{bbm}
\usepackage{nicefrac}       
\usepackage{microtype}      

\usepackage{amsfonts}       
\usepackage{amsthm}
\usepackage{amsmath}
\usepackage{amssymb}
\usepackage{mathtools}

\usepackage{graphicx}
\usepackage{booktabs}
\usepackage{array}
\usepackage{multirow}
\usepackage{adjustbox}
\usepackage{caption}
\usepackage{wrapfig}
\usepackage{placeins}

\usepackage{xcolor}         
\usepackage{xspace}
\usepackage{enumitem}
\usepackage{setspace}

\renewcommand{\ie}{\textit{i}.\textit{e}., }
\renewcommand{\eg}{\textit{e}.\textit{g}., }
\newcommand{\viz}{\textit{viz}., }

\def\rvx{{\mathbf{x}}}
\def\rvy{{\mathbf{y}}}
\def\rvz{{\mathbf{z}}}
\def\rveps{{\boldsymbol{\epsilon}}}

\newtheorem{theorem}{Theorem}

\newtheorem{lemma}[theorem]{Lemma}

\usepackage{color}
\definecolor{Blue9}{rgb}{0.098,0.3,0.9}
\definecolor{Red7}{rgb}{0.941, 0.243, 0.243}
\definecolor{Green7}{RGB}{55, 178, 77}
\definecolor{BrickRed}{rgb}{0.6,0,0}
\definecolor{RoyalBlue}{rgb}{0,0,0.8}
\definecolor{Tdgreen}{rgb}{0,0.4,0.7}

\definecolor{pinegreen}{rgb}{0.0, 0.47, 0.44}
\definecolor{cornellred}{rgb}{0.7, 0.11, 0.11}
\definecolor{cadmiumgreen}{rgb}{0.0, 0.42, 0.24}
\definecolor{spirodiscoball}{rgb}{0.06, 0.75, 0.99}
\definecolor{Gray}{gray}{0.9}

\usepackage{pifont}
\newcommand{\cmark}{\ding{51}}%
\newcommand{\xmark}{\ding{55}}%
\newcommand{\ck}{\color{Green7}{\cmark}}
\newcommand{\xk}{\color{Red7}{\xmark}}
\newcommand{\gua}{\color{Green7}{\uparrow}}
\newcommand{\rda}{\color{Red7}{\downarrow}}

\newcommand{\dif}[1]{\tiny{\textcolor{Green7}{$\boldsymbol{(#1)}$}}}
\newcommand{\gdif}[1]{\phantom{\tiny{$\boldsymbol{(#1)}$}}}

\newcommand{\method}{AENIB}

\newcommand{\citet}{\cite}
\newcommand{\citep}{\cite}

\usepackage[pagebackref,breaklinks,colorlinks]{hyperref}
\hypersetup{citecolor=Tdgreen}

\usepackage[capitalize]{cleveref}
\crefname{section}{Sec.}{Secs.}
\Crefname{section}{Section}{Sections}
\Crefname{table}{Table}{Tables}
\crefname{table}{Tab.}{Tabs.}

\def\cvprPaperID{12271} 
\def\confName{CVPR}
\def\confYear{2023}

\begin{document}

\title{Enhancing Multiple Reliability Measures via \\ Nuisance-extended Information Bottleneck}

\author{Jongheon Jeong$^\dagger$ \quad  Sihyun Yu$^\dagger$ \quad  Hankook Lee$^\ddag$\thanks{Work done at KAIST.} \quad Jinwoo Shin$^\dagger$ \\
$^\dagger$Korea Advanced Institute of Science and Technology (KAIST) \quad $^\ddag$LG AI Research\\
{\tt\small \{jongheonj,sihyun.yu,jinwoos\}@kaist.ac.kr \quad hankook.lee@lgresearch.ai}
}
\maketitle

\begin{abstract}
In practical scenarios where training data is limited, many predictive signals in the data can be rather from some biases in data acquisition (\ie less generalizable), so that one cannot prevent a model from co-adapting on such (so-called) ``shortcut'' signals: this makes the model fragile in various distribution shifts. To bypass such failure modes, we consider an adversarial threat model under a mutual information constraint to cover a wider class of perturbations in training. This motivates us to extend the standard \emph{information bottleneck} to additionally model the \emph{nuisance information}. We propose an autoencoder-based training to implement the objective, as well as practical encoder designs to facilitate the proposed hybrid discriminative-generative training concerning both convolutional- and Transformer-based architectures. Our experimental results show that the proposed scheme improves robustness of learned representations (remarkably without using any domain-specific knowledge), with respect to multiple challenging reliability measures. For example, our model could advance the state-of-the-art on a recent challenging OBJECTS benchmark in novelty detection by $78.4\% \rightarrow 87.2\%$ in AUROC, while simultaneously enjoying improved corruption, background and (certified) adversarial robustness. Code is available at \url{https://github.com/jh-jeong/nuisance_ib}.
\end{abstract}

\begin{figure}[t]
  \centering
  \includegraphics[width=\linewidth]{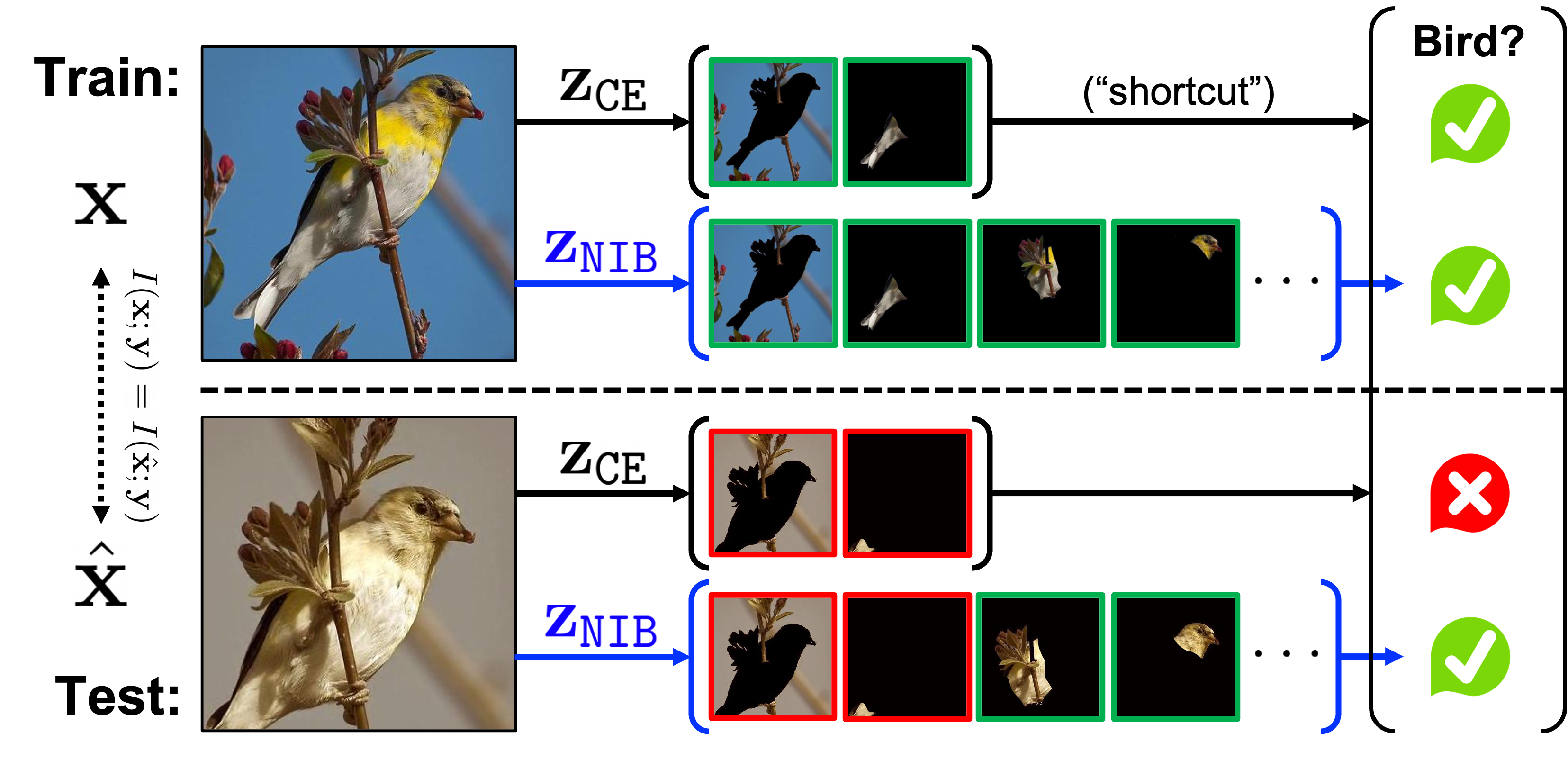}
  \caption{A high-level illustration of our method, \emph{nuisance-extended information bottleneck} (NIB). In this paper, we focus on scenarios when the input $\rvx$ can be corrupted $\rvx \rightarrow \hat{\rvx}$ in test-time while preserving its semantics. Unlike the standard cross-entropy training (CE), NIB aims to encode \emph{every} target-correlated signal in $\rvx$, some of which can be more reliable under distribution shifts.}
  \label{fig:teaser}
\vspace{-0.1in}
\end{figure}

\section{Introduction}
\label{s:intro}

Despite the recent breakthroughs in computer vision in aid of deep learning, \eg in image/video recognition~\cite{brock2021high,dai2021coatnet,tong2022videomae}, synthesis \cite{karras2020analyzing, rombach2022high,yu2022generating,ho2022imagen}, and 3D scene rendering~\cite{mildenhall2021nerf,tancik2022block, mildenhall2022dark}, deploying deep learning models to the real-world still places a burden on contents providers as it affects the \emph{reliability} of their services. In many cases, \emph{deep neural networks} make substantially fragile predictions for \emph{out-of-distribution} inputs, \ie samples that are not likely from the training distribution, even when the inputs are semantically close enough to in-distribution samples for humans \cite{szegedy2014intriguing,hendrycks2018benchmarking}. Such a vulnerability can be a significant threat in risk-sensitive systems, such as autonomous driving, medical imaging, and health-care applications, to name a few \cite{amodei2016concrete}. Overall, the phenomena highlight that deep neural networks tend to extract ``shortcut'' signals \cite{geirhos2020shortcut} from given limited (or potentially biased) training data in practice. 

To address such concerns, multiple literatures have been independently developed based on different aspects of model reliability. Namely, their methods use different \emph{threat models} and benchmarks, depending on how a shift in input distribution happens in test-time, and how to evaluate model performance against the shift. For example, in the context of \emph{adversarial robustness} \cite{madry2018towards,carlini2019evaluating,cohen2019certified,zhang2019theoretically}, a typical threat model is to consider the \emph{worst-case} noise inside a fixed $\ell_p$-ball around given test samples. Another example of \emph{corruption robustness} \cite{hendrycks2018benchmarking,hendrycks2020augmix,Hendrycks_2021_ICCV,wang2021tent} instead assumes pre-defined types of common corruptions (\eg Gaussian noise, fog, \emph{etc.}) that applies to the test samples. Lastly, \emph{novelty detection} \cite{hendrycks2017a,liang2018enhancing,lee2018maha,liu2020energy} usually tests whether a model can detect a specific benchmark dataset as out-of-distribution from the (in-distribution) test samples.

Due to discrepancy between each of ``ideal'' objectives and its practical threat models, however, the literatures have commonly found that optimizing under a certain threat model often hardly generalizes to other threat ones: \eg 
(a) several works \cite{xie2020adversarial,chun2020empirical,kireev2022effectiveness} have observed that standard adversarial training \cite{madry2018towards} often harms other reliability measures such as corruption robustness or uncertainty estimation, as well as its classification performance; 
(b) Hendrycks et al.~\citet{Hendrycks_2021_ICCV} criticize that none of the previous claims on corruption robustness could consistently generalize on a more comprehensive benchmark.
This also happens even for threat models targeting the same objective: \eg
(c) Yang et al.~\citet{yang2022fsood} show that state-of-the-arts in novelty detection are often too sensitive, so that they tend to also detect ``near-in-distribution'' samples as out-of-distribution and perform poorly on a benchmark regarding this. 
Overall, these observations suggest that one should avoid optimizing reliability measures assuming a specific threat model or benchmark, and motivate to find a new threat model that is generally applicable for diverse scenarios of reliability concerns.

\vspace{0.05in}
\noindent\textbf{Contribution. } In this paper, we propose \emph{nuisance-extended information bottleneck} (NIB), a new training objective targeting model reliability without assuming a prior on domain-specific tasks. Our method is motivated by rethinking the \emph{information bottleneck} (IB) principle \cite{tishby99information,naf} under presence of distribution shifts. Specifically, we argue that a ``robust'' representation $\rvz \coloneqq f(\rvx)$ should always encode \emph{every} signal in the input $\rvx$ that is correlated with the target $\rvy$, rather than extracting only a few shortcuts (\eg Figure~\ref{fig:teaser}). This motivates us to consider an \emph{adversarial} form of threat model on distribution shifts in $\rvx$, under a constraint on the mutual information $I(\rvx, \rvy)$. To implement this idea, we propose a practical design by incorporating a \emph{nuisance representation} $\rvz_n$ alongside $\rvz$ of the standard IB so that $(\rvz, \rvz_n)$ can reconstruct $\rvx$. This results in a novel synthesis of \emph{adversarial autoencoder} \cite{makhzani2015adversarial} and \emph{variational IB} \citep{alemi2016deep} into a single framework. For the architectural side, we propose (a) to utilize the \emph{internal feature statistics} for convolutional network based encoders, and (b) to incorporate \emph{vector-quantized} patch representations for Transformer-based \cite{dosovitskiy2021an} encoders to model $\rvz_n$, mainly to efficiently encode the nuisance $\rvz_n$ (as well as $\rvz$) in a scalable manner.

We perform an extensive evaluation on the representations learned by our scheme, showing comprehensive improvements in modern reliability metrics: including (a) novelty detection, (b) corruption (or natural) robustness, (c) background robustness and (d) certified adversarial robustness. The results are particularly remarkable as the gains are not from assuming a prior on each of specific threat models. For example, we obtain a significant reduction in \mbox{CIFAR-10-C} error rates of the highest severity, \ie by $26.5\% \rightarrow 19.5\%$, without extra {domain-specific} prior as assumed in recent methods \cite{hendrycks2020augmix,hendrycks2022pixmix}. Here, we also show that the effectiveness of our method is scalable to larger-scale (ImageNet) datasets. 
For novelty detection, we could advance AUROCs in recent OBJECTS \cite{yang2022fsood} benchmarks by a large margin of $78.4\% \rightarrow 87.2\%$ in average upon the state-of-the-art, showing that
our representations can provide a more semantic information to better discriminate out-of-distribution samples. Finally, we also demonstrate how the representations can further offer enhanced robustness against adversarial examples, by applying randomized smoothing \cite{cohen2019certified} on them.

\begin{figure*}[t]
  \centering
  \includegraphics[width=0.9\linewidth]{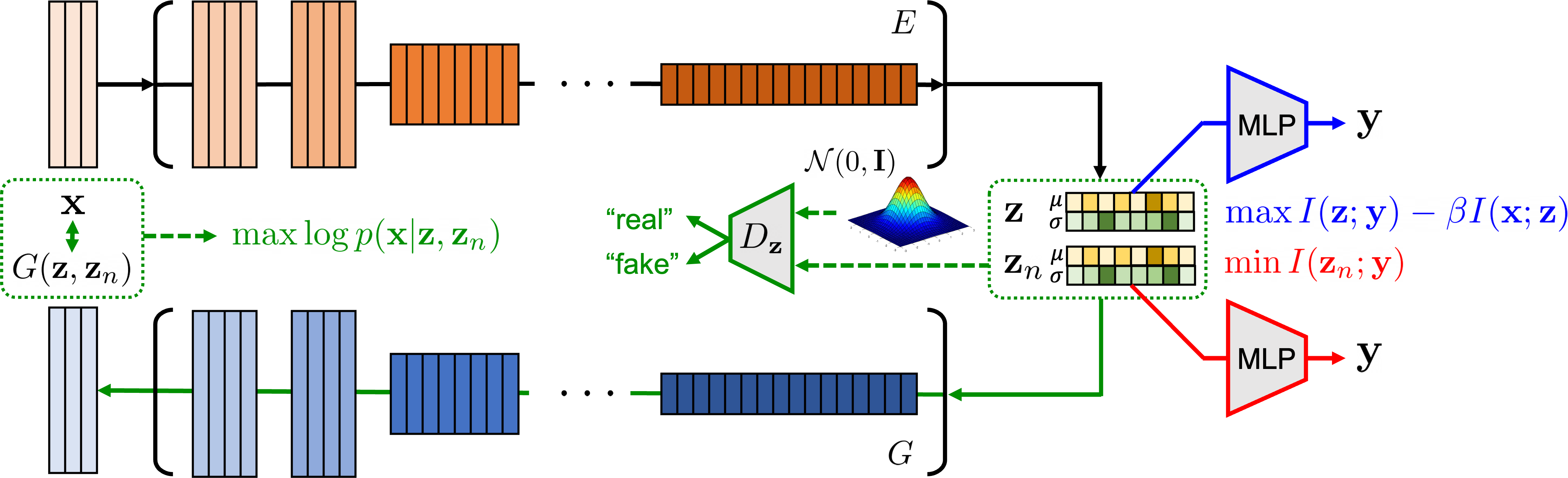}
  \caption{An overview of the proposed framework, the \emph{autoencoder-based nuisance-extended information bottleneck} (AENIB). 
  It illustrates the general pipeline, and Appendix~\ref{appendix:architecture} provides specific instantiations for convolutional and Transformer-based architectures. Overall, we incorporate \textcolor{Green7}{adversarial autoencoder} into \textcolor{blue}{variational information bottleneck} by introducing a \textcolor{red}{\emph{nuisance} $\rvz_n$ (to $\rvy$)} in representation learning.}
  \label{fig:overall}
\vspace{-0.1in}
\end{figure*}

\section{Background}

\vspace{0.05in}
\noindent\textbf{Notation. } Given two random variables $\rvx \in \mathcal{X}$, the input, and $\rvy \in \mathcal{Y}$, the target, 
our goal is is to find a mapping (or an \emph{encoder}) $f: \mathcal{X} \rightarrow \mathcal{Z}$ from data $\mathcal{D}=\{(x_i, y_i)\}^n_{i=1}$\footnote{Although we focus on \emph{supervised learning}, the framework itself in general does not rule out more general scenarios, \eg when the target $\rvy$ can be \emph{self-supervised} from $\rvx$ \cite{oord2018representation,chen2020simclr}.} so that $\rvz:=f(\rvx)$, the representation, can predict $\rvy$ with a simper (\eg linear) mapping \cite{bell1995information,kohonen1990self}. We assume that $f$ is parametrized by a neural network, and is \emph{stochastic} to adopt an information theoretic view \cite{naf}, \ie the encoder output is a random variable defined as $p_f(\rvz|\rvx)$ rather than a constant. Such a modeling can be done through the \emph{reparametrization trick} \cite{kingma2014autoencoding} with an independent random variable $\rveps$ and (deterministic) $f$ by assuming $\rvz := f(\rvx, \rveps)$. For example, one of standard designs parametrizes $f$ by:
\begin{equation}\label{eq:gaussian_decoder}
    f(\rvx, \rveps):= f^{\mu}(\rvx) + \rveps \cdot  f^{\sigma}(\rvx),
\end{equation}
{where $f^{\mu}\in\mathbb{R}^{|\mathcal{Z}|}$ and $f^{\sigma}\in\mathbb{R}_{+}^{|\mathcal{Z}|}$ are deterministic mappings modeling $\mu$ and $\sigma$ in $\mathcal{N}(\rvx; \mu, \sigma^2I)$, respectively, 
so that they can still be learned through a gradient-based optimization.}

The data $\mathcal{D}$ is usually assumed to consist of \textit{i.i.d.}\ samples from a certain \emph{data generating distribution} $(x_i, y_i) \sim p_d(\rvx, \rvy)$. One expects that $f$ learned from $\mathcal{D}$ could generalize to predict $p_d(\rvy|\rvx)$ for unseen samples from $p_d(\rvx, \rvy)$. The formulation, however, does not specify how $f$ should {behave} for inputs that are not likely from $p_d$, say $\hat{x}$. This becomes problematic for those who expect that the decision making of $f$ should be close to that of human being, at least when $\hat{x}$ differs from $p_d$ only up to what humans regard as \emph{nuisance}. This is where the current neural networks commonly fail under the standard training practices. 

\vspace{0.05in}
\noindent\textbf{{Information bottleneck. }} Intuitively, a ``good'' representation $\rvz$ should keep information of $\rvx$ that is correlated with $\rvy$, while preventing $\rvz$ from being too complex. The \emph{information bottleneck} \cite{tishby99information,naf} (IB) is a principled approach to obtain such a succinct representation $\mathbf{x} \rightarrow \mathbf{z}$ for a given downstream task $\mathbf{x} \rightarrow \mathbf{y}$: namely, it finds $\mathbf{z}$ that (a) maximizes the (task-relevant) mutual information $I(\mathbf{z}; \mathbf{y})$, while (b) minimizing $I(\mathbf{x}; \mathbf{z})$ to constrain the capacity of $\mathbf{z}$ for better generalization. 
In other words, it sets the \emph{mutual information} $I(\rvx;\rvz)$ as the complexity measure of~$\rvz$. Specifically, it aims to maximize the following objective: 
\begin{equation}\label{eq:ib}
    \max_f R_{\tt IB}(f),\quad\mbox{for}\quad R_{\tt IB}(f) := I(\rvz;\rvy) - \beta I(\rvx;\rvz), 
\end{equation}
where $\beta \ge 0$ controls the capacity constraint which ensures $I(\rvx;\rvz) \le I_\beta$ for some $I_\beta$. 

\section{Nuisance-extended information bottleneck}

The standard information bottleneck (IB) objective \eqref{eq:ib} obtains a representation $\rvz:=f(\rvx)$ on premise that the future inputs will be also from $p_d(\rvx, \rvy)$. In this paper, we aim to extend IB under assumption that the input $\rvx$ can be corrupted through an \emph{unknown} noisy channel in the future, say $\rvx \rightarrow \hat{\rvx}$, while $\hat{\rvx}$ still preserves the \emph{semantics} of $\rvx$ with respect to $\rvy$: in other words, we assume $I(\rvx; \rvy) = I(\hat{\rvx}; \rvy) > 0$. Intuitively, one can imagine a scenario that $\rvx$ contains multiple signals that each is already highly correlated with $\rvy$, \ie filtering out the remainder from $\rvx$ does not affect its mutual information with $\rvy$. It may or may not be surprising that such signals are quite prevalent in deep neural networks, \eg \citet{ilyas2019adversarial} empirically observe that {adversarial perturbations} \cite{szegedy2014intriguing,goodfellow2014explaining} are {sufficient} for a model to perform accurate classification.

In the context of IB framework, where the goal is to obtain a succinct encoder $f$, it is now reasonable to presume that the noisy channel $\hat{\rvx}$ acts like an \emph{adversary}, \ie it minimizes:
\begin{equation}\label{eq:x_hat}
    \min_{\hat{\rvx} }I(\hat{\rvz}:=f(\hat{\rvx}); \rvy) \text{\ \ subject to\ \ } I(\rvx; \rvy) = I(\hat{\rvx}; \rvy),
\end{equation}
given that one has no information on how the channel would behave \emph{a priori}. This optimization thus would require $f$ to extract \emph{every} signal in $\rvx$ whenever it is highly correlated with $\rvy$, to avoid the case when $\hat{\rvx}$ filters out all the signal except one that $f$ has missed. We notice that, nevertheless, directly optimizing \eqref{eq:x_hat} with respect to $\hat{\rvx}$ is computationally infeasible in practice, considering that (a) it is in many cases an unconstrained optimization in a high-dimensional $\mathcal{X}$, (b) with a constraint on (hard-to-compute) mutual information. 

In this paper, to make sure that $f$ still exhibits the adversarial behavior without \eqref{eq:x_hat}, we propose to let $f$ to model the \emph{nuisance representation} $\rvz_n$ as well as $\rvz$. Specifically, $\rvz_n$ aims to model the ``remainder'' information from $\rvz$ needed to reconstruct $\rvx$, \ie it maximizes $I(\rvx;\rvz, \rvz_n)$. At the same time, $\rvz_n$ compresses out any information that is correlated with $\rvy$, \ie it also minimizes $I(\rvz_n; \rvy)$. Therefore, every information that is correlated with $\rvy$ should be encoded into $\rvz$ in a complementary manner. Here, we remark that now the role of the capacity constraint in \eqref{eq:ib} becomes more important: not only for regularizing $\rvz$ to be simpler, it also penalizes $\rvz_n$ from pushing out unnecessary information to predict $\rvy$ into $\rvz$, making the objective competitive again between $\rvz$ and $\rvz_n$ as like in \eqref{eq:x_hat}. Combined with the original IB \eqref{eq:ib}, we {define \emph{nuisance-extended IB} (NIB) as the following}: 
\begin{equation}\label{eq:nib}
    \max_f R_{\tt NIB}(f) := {R_{\tt IB}(f)} - I(\rvz_n;\rvy) + \alpha I(\rvx; \rvz, \rvz_n), 
\end{equation}
where $\alpha \ge 0$. 
{The proposed {NIB} objective can be viewed as a regularized form of IB by introducing {a nuisance} $\rvz_n$. {Specifically, this optimization additionally forces $I(\mathbf{x}; \mathbf{z}, \mathbf{z}_n)$ and $I(\mathbf{z}_n;\mathbf{y})$ in \eqref{eq:nib} to be maximized and minimized, \ie $H({\mathbf{x}}|{\mathbf{z}},{\mathbf{z}}_n)=0$ and $I({\mathbf{z}}_n; \mathbf{y}) = 0$, respectively.} The following highlights that having these conditions, also with the independence $\rvz\perp\rvz_n$, leads $f$ that can recover the original $I(\rvx;\rvy)$ from $I(\hat{\rvz}; \rvy)$ (see Appendix~\ref{appendix:proof} for the derivation):
\begin{lemma}\label{thm:noisy}
Let $\rvx \in \mathcal{X}$, and $\rvy \in \mathcal{Y}$ be random variables, $\hat{\rvx}$ be a noisy observation of $\rvx$ with $I(\rvx;\rvy) = I(\hat{\rvx};\rvy)$. Given that $[\hat{\rvz}, \hat{\rvz}_n] := f(\hat{\rvx})$ of $\hat{\rvx}$ satisfies (a) $H(\hat{\rvx}|\hat{\rvz}, \hat{\rvz}_n)=0$, (b) $I(\hat{\rvz}_n; \rvy) = 0$, and (c) $\hat{\rvz} \perp \hat{\rvz}_n$, it holds $I(\hat{\rvz};\rvy) = I(\rvx;\rvy)$. 
\end{lemma}
}

In the following sections, we provide a practical design of the proposed NIB based on an autoencoder-based architecture. {Section~\ref{ss:training} and \ref{ss:arch} detail out its losses and architectures, respectively, and Section~\ref{ss:overall} summarizes the overall training.} Figure~\ref{fig:overall} illustrates an overview of our framework.

\subsection{{{\method}: A practical autoencoder-based design}}
\label{ss:training}

{Based on the NIB objective defined in \eqref{eq:nib} {and Lemma~\ref{thm:noisy}},} we design a practical training objective to implement the proposed framework. {Here, we present a simple instantiation of NIB with an autoencoder-based architecture upon \emph{variational information bottleneck} (VIB) \cite{alemi2016deep}, calling it {\emph{autoencoder-based NIB}} ({\method}).}

Overall, Lemma~\ref{thm:noisy} states that a robust encoder $f$ demands for a ``good'' nuisance model that achieves generalization on $\hat{\rvz}$ in three aspects: (a) a \emph{good reconstruction}, (b) \emph{nuisance-ness}, and (c) the \emph{independence between $\rvz$ and $\rvz_n$}. 
{To model these behaviors, we consider a decoder $g: \mathcal{Z} \rightarrow \mathcal{X}$ as well as the encoder $f: \mathcal{X} \rightarrow  \mathcal{Z}$, and adopt the following practical training objectives which incorporates an autoencoder-based loss and two adversarial losses \cite{goodfellow2014gan}:}
\begin{enumerate}[label=(\alph*),leftmargin=8mm]
    \item {We first pose a reconstruction loss to maximize $\log p(\rvx|\rvz, \rvz_n)$; standard designs assume the decoder output to follow $\mathcal{N}(\rvx, \sigma^2 \mathbf{I})$, which is equivalent to the \emph{mean-squared error} (MSE). Here, we use the \emph{normalized MSE} to efficiently balance with other losses}:\footnote{{We also explore a SSIM-based \cite{wang2004image} reconstruction loss as given in Appendix~\ref{appendix:architecture}, which we found beneficial for robustness particularly with Transformer-based models.}}
    \begin{equation}\label{eq:recon}
    \noindent    L_{\tt recon} := \tfrac{1}{\|\rvx\|_2^2}\|\rvx - {g}(\rvz, \rvz_n) \|_2^2 
    \end{equation}
    \item To force the nuisance-ness of $\rvz_n$ with respect to $\rvy$, we approximate $p(\rvy|\rvz_n)$ with a multi-layer perceptron (MLP), say $q_{n}$, and perform an adversarial training: 
    \begin{multline}\label{eq:nuis}
        L_{\tt nuis} := \mathbb{E}_{\rvx}[\mathbb{CE}(q^*_n(\rvz_n), \tfrac{\bm{1}}{|\mathcal{Y}|})], \\ \text{where } q^*_{n} := \min_{q_{n}} \mathbb{E}_{\rvx, \rvy}[\mathbb{CE}(q_n(\rvz_n), \rvy)],
    \end{multline}
    and $\mathbb{CE}$ denotes the cross entropy loss. Here, it optimizes $\mathbb{CE}$ towards the uniform distribution in $\mathcal{Y}$.\footnote{Alternatively, one can directly maximize $\mathbb{CE}(q^*_n(\rvz_n), \rvy)$; we use the current design to avoid potential instability of the maximization-based loss.}
    \item To induce the independence between $\rvz$ and $\rvz_n$, we assume that the joint prior of $\rvz$ and $\rvz_n$ is the isotropic Gaussian, \ie $p(\rvz, \rvz_n) \sim \mathcal{N}(0, \mathbf{I})$, and performs a GAN training with a 2-layer MLP discriminator $q_{\rvz}$:
    \begin{multline}\label{eq:loss_ind}
        L_{\tt ind} := \max_{q_{\rvz}} \mathbb{E}_{\rvx}[\log (q_{\rvz}({f}(\rvx)))] \\
        + \mathbb{E}_{\rvz, \rvz_n \sim \mathcal{N}(0, \mathbf{I})}[\log (1 - q_{\rvz}(\rvz, \rvz_n))].
    \end{multline}
\end{enumerate}
Lastly, to approximate the original IB objective $R_{\tt IB}(f)$ in NIB \eqref{eq:nib}, we instead maximize the \emph{variational information bottleneck} (VIB) \cite{alemi2016deep} {objective} $L_{\tt VIB}^{\beta}$, that can provide a lower bound on $R_{\tt IB}$.\footnote{{A more detailed description on the VIB framework (as well as on GAN) can be found in Appendix~\ref{appendix:tech}.}} Specifically, it makes {variational} approximations of: (a) $p(\rvy|\rvz)$ by a (parametrized) decoder neural network $q(\rvy|\rvz)$, and (b) $p(\rvz)$ by an ``easier'' distribution $r(\rvz)$, \eg isotropic Gaussian $\mathcal{N}(\rvz|0, \mathbf{I})$. Assuming a Gaussian decoder \eqref{eq:gaussian_decoder} for $f(\rvx, \rveps)$, we have:
\begin{multline}\label{eq:loss_vib}
    L_{\tt VIB}^{\beta} := \frac{1}{n} \sum_{i=1}^{n}\ \mathbb{E}_{\rveps}[-\log q(y_i | f(x_i, \rveps))] \\ + \beta\  \mathrm{KL}~(p(\rvz|x_i) \| r(\rvz)).
\end{multline}

\begin{table*}[t]
\centering
\small
    \begin{adjustbox}{width=0.9\linewidth}
    \begin{tabular}{l cccccc|c|cccccc|c}
    \toprule
     & \multicolumn{7}{c}{CIFAR-10-C} & \multicolumn{7}{c}{CIFAR-100-C} \\
    \cmidrule(r){2-8} \cmidrule(r){9-15}
         \multicolumn{1}{r}{Severity} & Clean & 1 & 2 & 3 & 4 & 5 & Avg. & Clean & 1 & 2 & 3 & 4 & 5 & Avg. \\
    \midrule
    {Cross-entropy}  & {6.08} & {8.89} & {11.1} & {14.0}	& {19.7}	& {26.5} & {16.0} & {25.1} & {31.4} & {35.1} & {39.3} & {46.8} & {54.0} & {41.3} \\
     {VIB} \cite{alemi2016deep} 
    & {5.98} &  {8.68}	& {10.7} & {13.4} & {18.6} & {24.9} & {15.2} & {26.0} & {31.9} & {35.9} & {40.4} & {47.8} & {55.2} & {42.2} \\
     {AugMix} \cite{hendrycks2020augmix} 
    & {6.52} & {8.97} & {10.8} & {13.4} & {18.4}	& {23.9} & {15.1} & {24.9} & {29.9} & {33.3} & {37.1} & {43.6} & {51.1} & {39.0} \\
     {PixMix}$^\dagger$  \cite{hendrycks2022pixmix} 
    & {5.43} & {\underline{7.10}} & {\underline{8.14}} & {\underline{9.40}} & {\underline{12.1}} & {\underline{14.9}} & {\underline{10.3}} & {23.2} & 26.7 & \underline{28.7} & \underline{30.8} & \underline{35.0} & \underline{39.0} & \underline{32.0} \\
    \cmidrule(r){1-1} \cmidrule{2-2} \cmidrule(lr){3-8} \cmidrule(lr){9-9} \cmidrule(l){10-15}
    {\textbf{{\method} (ours)}} & {\underline{4.97}} & {{7.49}} & {{8.96}} & {{11.0}} & {14.8} & {19.5}
    & {12.3} & {{22.6}} & {{27.6}} & {{30.5}} & {{34.1}} & {{39.8}} & {{47.1}} & {{35.8}} \\
     {{+ AugMix}} \cite{hendrycks2020augmix} 
    & {5.35} & {7.65} & {8.99} & {{11.0}} & {{14.2}} & {{18.4}}
    & {{12.0}} & {\underline{21.9}} & {\underline{26.4}} & {{29.1}} & {{32.4}} & {{37.8}} & {{44.3}} & {{34.0}} \\
     {{+ PixMix}}$^\dagger$ \cite{hendrycks2022pixmix} 
     & {\textbf{4.67}} & {\textbf{5.90}} & {\textbf{6.55}} & {\textbf{7.45}} & {\textbf{9.12}} & {\textbf{11.4}} & {\textbf{8.08}} & 
    \textbf{21.2} & \textbf{24.4} & \textbf{26.0} & \textbf{27.8} & \textbf{31.1} & \textbf{34.8} & \textbf{28.8} \\
    \bottomrule
\end{tabular}
    \end{adjustbox}
    \vspace{-0.05in}
    \caption{Comparison of average corruption error rates (\%; $\rda$) per severity level on \mbox{CIFAR-10/100-C} \cite{hendrycks2018benchmarking}. Bold and underline denote the best and runner-up, respectively.
    $^\dagger$PixMix \cite{hendrycks2022pixmix} utilizes an external dataset consisting of pattern- and fractal-like images.}\label{tab:corruption}
    \vspace{-0.1in}
\end{table*}

\begin{table}[t]
\centering
\small
    \begin{adjustbox}{width=0.85\linewidth}
    \begin{tabular}{l ccccc}
\toprule
Method & C10 & C10-C & C10.1  & C10.2  & CINIC  \\
\midrule
Cross-entropy & 6.08 & 16.0 & 13.4 & 18.3 & 23.7 \\
VIB \cite{alemi2016deep} 
& 5.98 & 15.2 & 13.6 & 16.8 & 23.6 \\
NLIB \cite{kolchinsky2019nonlinear} & 6.81 & 17.0 & 14.6 & 17.5 & 24.3 \\
sq-NLIB \cite{thobaben2020convex} & 6.02 & 15.5 & 13.0 & 17.1 & 23.7 \\
DisenIB \cite{pan2021disentangled} & 5.76 & 15.2 & 13.2 & 17.2 & 23.7 \\
\midrule
{AugMix \cite{hendrycks2020augmix}} 
& 6.52 & 15.1 & 14.2 & 17.2 & 24.2 \\
{PixMix \cite{hendrycks2022pixmix}} 
& 5.43 & \underline{10.3} & 13.1 & 16.6 & 23.2 \\
\midrule
{\textbf{{\method} (ours)}} & \underline{4.97} & {12.3} & \underline{11.6} & \underline{15.5} & \underline{22.2} \\
{+ AugMix \cite{hendrycks2020augmix}} 
& 5.35 & 12.0 & 12.5 & 15.8 & 22.6\\
{+ PixMix \cite{hendrycks2022pixmix}} 
& \textbf{4.67} & \textbf{8.08} & \textbf{10.4} & \textbf{14.8} & \textbf{22.1} \\
\bottomrule
\end{tabular}

    \end{adjustbox}
    \vspace{-0.05in}
    \caption{{Comparison of test error rates (\%;~$\rda$) on CIFAR-10 and its variants: CIFAR-10-C/10.1/10.2, and CINIC. Bold and underline indicate the best and runner-up results, respectively.}}\label{tab:corruption_vit}
    \vspace{-0.1in}
\end{table}

\subsection{{Architectures for nuisance modeling}}
\label{ss:arch}

In principle, our framework is generally compatible with any encoder architectures: \eg say an encoder $f: \mathcal{X} \rightarrow  \mathcal{Z}$ and decoder $g: \mathcal{Z} \rightarrow \mathcal{X}$, respectively. In order to apply VIB, we assume that the encoder has two output heads of dimension $2K$, where $K$ denotes the dimension of $\rvz$. Here, each output head models a Gaussian random variable by reparametrization, \ie by modeling $(\mu, \sigma)$ as the encoder output for both $\rvz \in \mathbb{R}^K$ and $\rvz_n\in \mathbb{R}^{K_n}$. 

Although it is possible that $f$ models $\rvz$ and $\rvz_n$ by simply taking deep feed-forward representations following conventions, we observe that modeling nuisances $\rvz_n$ (which is essentially ``generative'') in standard (discriminative) architectures incur a training instability thus in performance: the nuisance information often requires to model finer details in a given inputs, which may be available rather in early layers of $f$, but not in the later layers for classification. 

In this paper, we propose simple architectural treatments to improve the stability of nuisance modeling concerning both convolutional networks and Vision Transformers~\cite{dosovitskiy2021an} (ViTs). 
This section focuses on introducing the design for convolutional networks, and we refer the readers for the ViT-based design to Appendix~\ref{appendix:architecture}: which is even simpler thanks to their patch-level representations available.

Given a convolutional encoder $f$,  we encode $\rvz_n$ (as well as $\rvz$) from the collection of \emph{internal features statistics}, rather than directly using the output of $f$.
Specifically, we extract $L$ intermediate feature maps of a given input $\rvx$, namely $\rvx^{(1)}, \cdots, \rvx^{(L)}$ from $f(\rvx)$, and define the \emph{projection} of $\rvx$ by:
\begin{equation}\label{eq:projection}
    \Pi_{f}(\rvx) := \begin{bmatrix}
        \mathbf{m}^{(1)} & \mathbf{m}^{(2)} & \cdots & \mathbf{m}^{(L)} \\
        \mathbf{s}^{(1)} & \mathbf{s}^{(2)} & \cdots & \mathbf{s}^{(L)}
    \end{bmatrix},
\end{equation}
where $\mathbf{m}^{(l)}$ and $\mathbf{s}^{(l)}$ are the first and second moment of feature maps in $\rvx^{(l)}$, assuming that $\rvx^{(l)} \in \mathbb{R}^{HWC}$: 
$\mathbf{m}^{(l)}_c \coloneqq \tfrac{1}{HW}\sum_{hw} \rvx^{(l)}_{hwc}, \text{ and } \mathbf{s}^{(l)}_c \coloneqq \tfrac{1}{HW}\sum_{hw} (\rvx^{(l)}_{hwc} - \mathbf{m}^{(l)}_c)^2$.

{In Appendix~\ref{appendix:generation}, we demonstrate that this simple projection can sufficiently encode a \emph{generative} representation of $\rvx$: \viz we show that one can successfully and efficiently train GANs with a discriminator defined upon $\Pi_f$.}
Motivated by this observation, we adopt $\Pi_f$ in modeling the encoder representations $\rvz$ and $\rvz_n$. We encode $\rvz$ and $\rvz_n$ by simply applying MLPs to $\Pi_f(\rvx)$ \eqref{eq:projection}. Despite its simplicity, we observe this treatment enables a stable training of {\method}.

\subsection{{Overall training objective}}
\label{ss:overall}

Combining the proposed objectives as well as the VIB loss, $L_{\tt VIB}^{\beta}$ \eqref{eq:loss_vib} leads us to the final objective. Although combining multiple losses in practice may introduce additional hyperparameters, we found most of the proposed losses can be added without scaling except for the reconstruction loss $L_{\tt recon}$ and the $\beta$ in the original VIB loss. Hence, we get: 
\begin{equation}\label{eq:final}
    L_{\tt {\method}} := L_{\tt VIB}^{\beta} + \alpha \cdot L_{\tt recon} + L_{\tt nuis} + L_{\tt ind}.
\end{equation}
Algorithm~\ref{alg:training} in Appendix~\ref{appendix:alg} summarizes the procedure.

\section{Experiments}
\label{s:experiments}

We verify the effectiveness of our proposed {{\method} training} for various aspects of model reliability: specifically, we cover (a) corruption and natural robustness (Section~\ref{exp:corruption}), (b) novelty detection (Section~\ref{exp:ood}), {and} (c) certified adversarial robustness (Section~\ref{exp:adv}) tasks which all have been challenging without task-specific priors \cite{hendrycks2020augmix,hendrycks2018deep,madry2018towards}. We provide an ablation study in Appendix~\ref{appendix:ablation} for a component-wise analysis. We also present an evaluation on our proposed components in the context of generative modeling in Appendix~\ref{appendix:generation}. The full details on the experiments, \eg datasets, training details, and hyperparameters, can be found in Appendix~\ref{appendix:setup}.

\subsection{{Robustness against natural corruptions}}\label{exp:corruption}

We first evaluate corruption robustness of our method, \ie its generalization ability under natural corruptions (\eg fog, brightness, \textit{etc}.) and distribution shifts those are still semantic to humans. To this end, {we consider a wide range of benchmarks those are derived from CIFAR-10 and ImageNet for the purpose of measuring generalization. Namely, for CIFAR-10 models we test on} (a) CIFAR-10/100-C~\citep{hendrycks2018benchmarking}, a corrupted version of CIFAR-10/100 simulating 15 common corruptions in 5 severity levels, as well as (b) CIFAR-10.1 \cite{recht2018cifar10.1}, CIFAR-10.2 \cite{lu2020harder}, and CINIC-10 \cite{darlow2018cinic}, \ie three re-generations of the CIFAR-10 test set. For ImageNet models, on the other hand, we test (a) ImageNet-C \citep{hendrycks2018benchmarking}, a corrupted version of ImageNet validation set, (b) \mbox{ImageNet-R}~\cite{Hendrycks_2021_ICCV}, a collection of rendition images for 200 ImageNet classes, and ImageNet-Sketch \cite{wang2019learning}, as well as (c) the Background Challenge \cite{xiao2021noise} benchmark to evaluate model bias against background changes. This section mainly reports the results from ViT \cite{dosovitskiy2021an,touvron2021training} based architectures, but we also report the results with ResNet-18 \cite{he2016deep} in Appendix~\ref{appendix:additional_exp}.

\begin{figure}[t]
\begin{minipage}{.47\linewidth}
    \centering
    \includegraphics[width=0.92\linewidth]{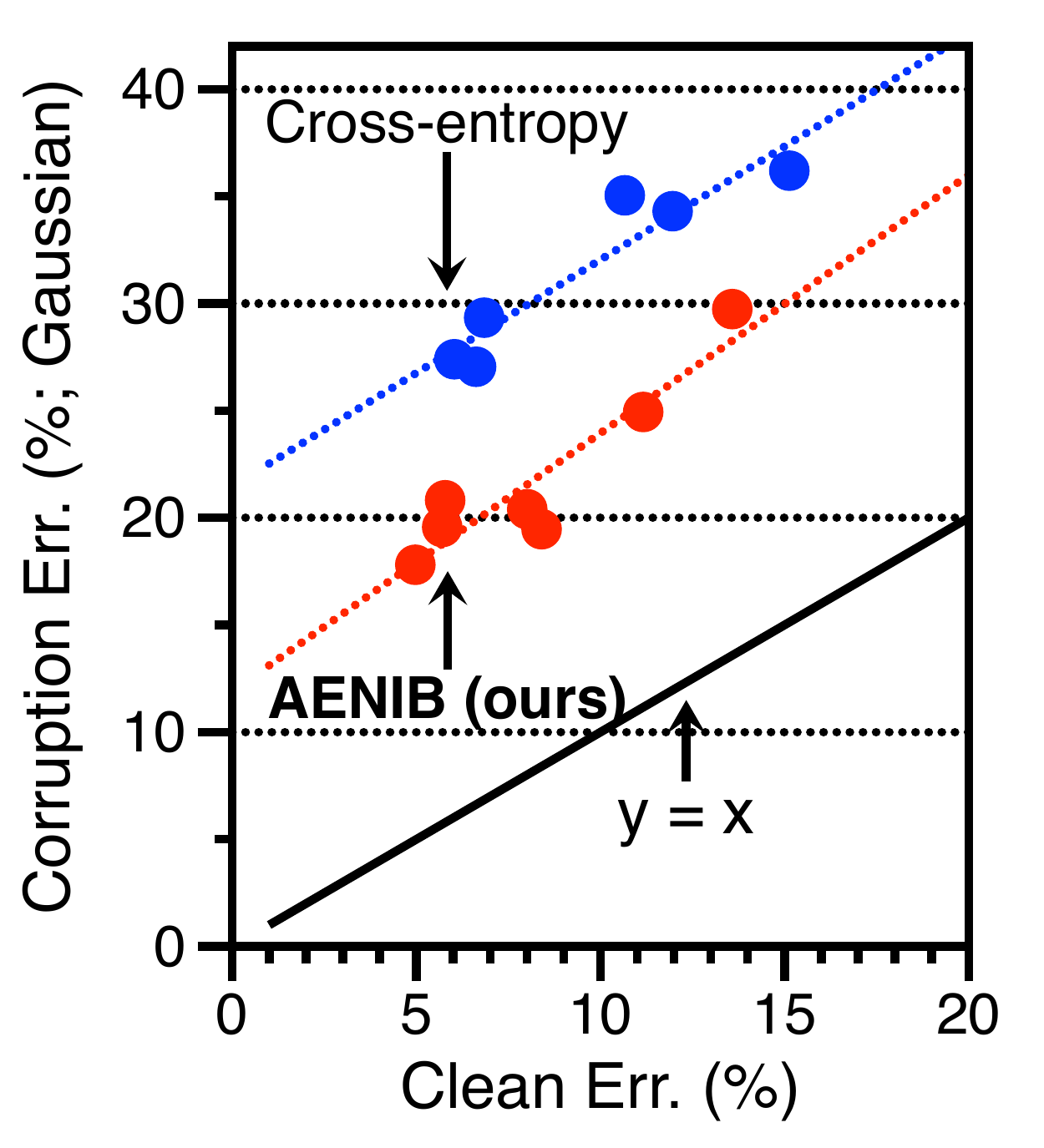}
    \vspace{-0.1in}
    \caption{{Comparison of trends in clean vs.~corruption errors against Gaussian on ViT-S/4.
    }}\label{figure:in-vs-out-error}
\end{minipage}
\hfill
\begin{minipage}{.47\linewidth}
    \centering
    \includegraphics[width=0.92\linewidth]{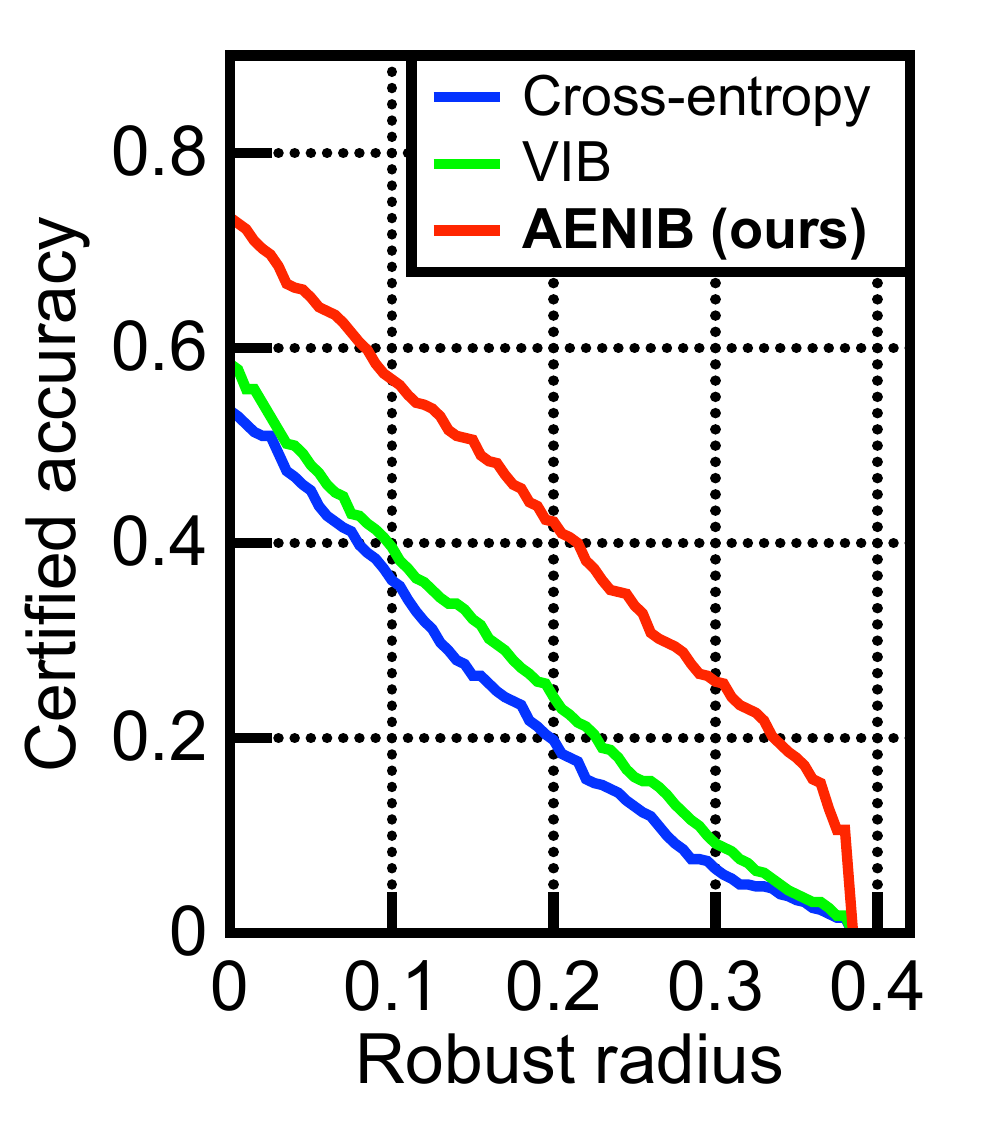}
    \vspace{-0.1in}
    \caption{Comparison of certified adversarial robust accuracy at various radii on CIFAR-10.}\label{figure:adv_robustness}
\end{minipage}
\vspace{-0.15in}
\end{figure}

{Table~\ref{tab:corruption} and \ref{tab:corruption_vit} summarize the results on CIFAR-based models.}
{In Table~\ref{tab:corruption}, we observe that {{\method}} significantly and consistently improves corruption errors upon VIB, and these gains are strong even compared with state-of-the-art methods: \eg 
{{\method}} can solely outperform a strong baseline of AugMix \cite{hendrycks2020augmix}. Although a more recent method of PixMix \cite{hendrycks2022pixmix} could achieve a lower corruption error by utilizing extra (pattern-like) data, we remark that (a) {{\method}} also benefit from PixMix (\ie the extra data) as given in ``{{\method}} + PixMix'', and (b) the results on Table~\ref{tab:corruption_vit} show that the generalization capability of {{\method}} is better than PixMix on CIFAR-10.1, 10.2 and CINIC-10, \ie in beyond common corruptions, by less relying on domain-specific data. 

{Next, Table~\ref{tab:imagenet} highlights that the effectiveness of {\method} can generalize to a more larger-scale, higher-resolution dataset of ImageNet: we still observe that {\method} can consistently improve robust accuracy for diverse corruption types, again without leveraging any further data augmentation during training.}
Figure~\ref{figure:in-vs-out-error} compares the linear trends made by Cross-entropy and {{\method}} across different data augmentations and hyperparameters, confirming that {{\method}} exhibits a better operating points even in terms of \emph{effective robustness} \cite{taori2020measuring}, given the recent observations on the correlation between in- \emph{vs.}~out-of-distribution performances across different models \cite{taori2020measuring,Hendrycks_2021_ICCV,miller2021accuracy}.} 

Lastly, Table~\ref{tab:background} further evaluates the ImageNet classifiers on \emph{Background Challenge} \cite{xiao2021noise}, a benchmark established to test the model robustness against background shifts: specifically, it constructs variants of ImageNet-9 (that combines 370 subclasses of ImageNet; \textsc{Original}) with different combinations of backgrounds. Our AENIB-based models still consistently improve upon the cross-entropy baseline on the benchmark, showing that AENIB indeed tends to learn less-biased features against background changes.  

\begin{table}[t]
\centering
\small
    \begin{adjustbox}{width=\linewidth}
    \begin{tabular}{lcccc}
\toprule
     & \multicolumn{2}{c}{ViT-S/16} & \multicolumn{2}{c}{ViT-B/16} \\
\cmidrule(r){2-3} \cmidrule(r){4-5} 
Dataset & Baseline & \textbf{AENIB (ours)} & Baseline & \textbf{AENIB (ours)} \\
\midrule
IN-1K & \textbf{25.1}  & \textbf{25.1} \gdif{-x.x}  & 21.8  & \textbf{21.9} \gdif{-x.x} \\
\midrule
IN-C (mCE) & 65.9  & \textbf{65.2} \dif{-0.7}  & 58.6  & \textbf{57.5} \dif{-1.1} \\
IN-R & 70.3  & \textbf{67.1} \dif{-3.2}  & 66.3  & \textbf{64.4} \dif{-1.9} \\
IN-Sketch  & 80.3  & \textbf{77.7} \dif{-2.6}  & 76.5 & \textbf{74.4} \dif{-2.1} \\
\bottomrule
\end{tabular}%
    \end{adjustbox}
    \vspace{-0.05in}
    \caption{Comparison of error rates (\%; $\rda$) or mean corruption error (mCE, \%; $\rda$) on ImageNet (IN) and its variants, namely IN-C \cite{hendrycks2018benchmarking}, IN-R \cite{Hendrycks_2021_ICCV}, and IN-Sketch \cite{wang2019learning}. Bold indicates the best results.}\label{tab:imagenet}
    \vspace{-0.1in}
\end{table}

\begin{table}[t]
\centering
\small
    \begin{adjustbox}{width=\linewidth}
    \begin{tabular}{lcccc}
\toprule
BG-Challenge      & \multicolumn{2}{c}{ViT-S/16} & \multicolumn{2}{c}{ViT-B/16} \\
\cmidrule(r){2-3} \cmidrule(r){4-5} 
Dataset & Baseline & \textbf{AENIB (ours)} & Baseline & \textbf{AENIB (ours)} \\
\midrule
\textsc{Original (IN-9; $\gua$)} & 95.3  & \textbf{95.5} \gdif{-x.x}  & 96.0  & \textbf{96.1} \gdif{-x.x} \\
\textsc{Only-BG-T} ($\rda$) & 20.3  & \textbf{17.8} \dif{-2.5}  & 24.2  & \textbf{21.1} \dif{-3.1} \\
\textsc{Mixed-Same} ($\gua$) & 86.3  & \textbf{88.3} \dif{+2.0}  & 87.4  & \textbf{88.9} \dif{+1.5} \\
\textsc{Mixed-Rand} ($\gua$) & 77.8  & \textbf{80.5} \dif{+2.7}  & 80.1  & \textbf{81.8} \dif{+0.7} \\
\midrule
BG-gap ($\rda$) & \phantom{x}8.5  & \phantom{x}\textbf{7.8} \dif{-0.7}  & \phantom{x}7.3  & \phantom{x}\textbf{7.1} \dif{-0.2} \\
\bottomrule
\end{tabular}%

    \end{adjustbox}
    \vspace{-0.05in}
    \caption{Evaluation of AENIB on Backgrounds Challenge \cite{xiao2021noise} compared to the cross-entropy baseline. All the models are trained on ImageNet, and warped to perform classification on ImageNet-9.}
    \label{tab:background}
    \vspace{-0.1in}
\end{table}

\begin{table}[t]
    \centering
    \Large
    \begin{adjustbox}{width=\linewidth}
    \begin{tabular}{llccccc}
    \toprule
    Method & Score &  SVHN & LSUN & ImageNet & C100 & {CelebA}   \\
    \midrule
    JEM \cite{Grathwohl2020Your}  
    & $\log p(x)$ & 0.67 & - & - & 0.67 & 0.75 \\
    JEM \cite{Grathwohl2020Your}  
    & $\max_y p(y|x)$ 
    \cite{hendrycks2017a} 
    & 0.89 & - & - & {0.87} & {0.79} \\
    SupCon \cite{khosla2020supervised} 
    & $\max_y p(y|x)$ \cite{hendrycks2017a}
    & 0.97 & 0.93 & 0.91 & \textbf{0.89} & - \\
    \midrule
    Cross-entropy  & $\max_y p(y|x)$ \cite{hendrycks2017a}
    & 0.94 & 0.94 & 0.92 & {0.86} & 0.64 \\
    Cross-entropy  & $\log \mathrm{Dir}_{0.05}(y)$ & 0.96 & 0.95 & 0.94 & {0.86} & 0.61 \\
    VIB \cite{alemi2016deep} 
    & $\max_y p(y|x)$ \cite{hendrycks2017a} 
    & 0.95 & 0.94 & 0.92 & {0.88} & 0.76 \\
    {VIB} \cite{alemi2016deep} 
    & {$\log \mathrm{Dir}_{0.05}(y)$} & {0.97} & {0.96} & {0.94} & {0.88} & {0.78} \\
    \midrule
   {\textbf{{\method} (ours)}} & {$\max_y p(y|x)$} \cite{hendrycks2017a} 
   & {0.88} & {0.88} & {0.86} & {0.84} & {\textbf{0.81}} \\
    {\textbf{{\method} (ours)}} & {$\log \mathrm{Dir}_{0.05}(y)$} & {0.90} & {0.95} & {0.92} & {0.86} & {0.80} \\
    {\textbf{{\method} (ours)}} & {$ + \log \mathcal{N}(z_n; 0, I)$} & {\textbf{0.98}} & {\textbf{0.99}} & {\textbf{0.99}} & {0.86} & {0.79} \\
    \bottomrule
\end{tabular}

    \end{adjustbox}
    \vspace{-0.05in}
    \caption{{Comparison of AUROC (\%; $\gua$) for OOD detection from CIFAR-10 with five OOD datasets: SVHN, LSUN, ImageNet, CIFAR-100, and CelebA. Bolds indicate the best results.}}\label{tab:ood}
    \vspace{-0.1in}
\end{table}

\begin{table*}[t]
\centering
\small
    \begin{adjustbox}{width=0.93\textwidth}
    \begin{tabular}{llcccc}
\toprule
 \multicolumn{2}{l}{FS-OOD: OBJECTS}
 & \multicolumn{4}{c}{AUROC (\%; $\gua$) / AUPR (\%; $\gua$) / FPR@TPR95 (\%; $\rda$)} \\
 \cmidrule{1-2} \cmidrule(l){3-6}
Method & Score & MNIST & FashionMNIST & {Texture} & CIFAR-100-C   \\
\midrule
Cross-entropy  & $\max_y p(y|x)$ \cite{hendrycks2017a}
& {66.98 / 52.66 / 93.54} & {73.78 / 90.15 / 88.08} & {74.18 / 93.34 / 85.64} & {74.12 / 89.74 / 87.26} \\
 & ODIN \cite{liang2018enhancing}
& {70.31 / 49.58 / 82.04} & {80.98 / 91.53 / \textbf{68.73}} & {70.14 / 89.97 / \underline{72.91}} & {67.51 / 83.97 / 84.26} \\
 & Energy-based \cite{liu2020energy}
& {54.55 / 34.14 / 92.23} & {76.50 / 89.80 / {72.40}} & {68.63 / 89.51 / 75.57} & {68.37 / 85.54 / 83.64} \\
 & Mahalanobis \cite{lee2018maha}
& {77.04 / 65.31 / 84.59} & {80.33 / 92.28 / 77.17} & {72.02 / 88.46 / 72.98} & {68.13 / 82.97 / 85.53} \\
 & SEM \cite{yang2022fsood}
& {75.69 / 76.61 / 99.70} & {79.40 / 93.14 / 93.72} & {\underline{79.69} / \underline{95.48} / 82.15} & {78.89 / 92.07 / 83.92} \\
\cmidrule{2-6}
 & $\boldsymbol{\log \mathrm{Dir}_{0.05}(y)}$ 
& {76.75 / 66.26 / 83.51} & {82.88 / 93.97 / 77.19} & {70.69 / 92.68 / 91.35} & {78.80 / 92.21 / 82.50} \\
\midrule
VIB \cite{alemi2016deep} 
& $\max_y p(y|x)$ \cite{hendrycks2017a}
& {80.23 / 73.50 / 80.69} & {76.35 / 91.22 / 84.75} & {74.67 / 94.09 / 87.22} & {76.12 / 91.03 / 84.99} \\
\cmidrule{2-6}
 & $\boldsymbol{\log \mathrm{Dir}_{0.05}(y)}$ 
& {86.13 / 79.45 / 64.92} & {81.11 / 93.12 / 77.82} & {73.84 / 93.50 / 88.00} & {78.54 / 91.85 / \underline{81.47}} \\
\midrule
{\textbf{{\method} (ours)}} & $\max_y p(y|x)$ \cite{hendrycks2017a} 
& {79.67 / 71.50 / 80.22} & {77.33 / 91.63 / 84.31} & {74.95 / 93.97 / 86.01} & {74.31 / 89.89 / 86.26} \\
\cmidrule{2-6}
 & $\boldsymbol{\log \mathrm{Dir}_{0.05}(y)}$   
& {\underline{90.53} / \underline{85.68} / \underline{52.08}} & {\underline{84.56} / \underline{94.61} / \underline{74.24}} & {75.04 / 93.83 / 86.01} & {\underline{79.39} / \underline{92.33} / 81.51} \\
 & $\boldsymbol{+\log \mathcal{N}(z_n; 0, I)}$   
& {\textbf{92.43} / \textbf{89.38} / \textbf{48.10}} & {\textbf{84.85} / \textbf{94.84} / {74.67}} & {\textbf{88.91} / \textbf{97.49} / \textbf{48.44}} & {\textbf{82.66} / \textbf{93.62} / \textbf{74.14}} \\
\bottomrule
\end{tabular}
    \end{adjustbox}
    \vspace{-0.05in}
    \caption{{Comparison of OOD detection performances on the OBJECTS benchmark \cite{yang2022fsood}, which considers CIFAR-10-C and ImageNet-10 as in-distribution as well as the training in-distribution of CIFAR-10. Bold and underline denote the best and runner-up results, respectively.}}\label{tab:newood_far}
    \vspace{-0.1in}
\end{table*}


\subsection{Novelty detection}
\label{exp:ood}

{Next, we show that our {{\method}} model can be also a good detector for \emph{out-of-distribution samples} (OODs), \ie to solve the \emph{novelty detection} task.} In general, the task is defined by a binary classification problem that aims to discriminate novel samples from in-distribution samples. A typical practice here is to define a \emph{score function} for each input, \eg the maximum confidence score \cite{hendrycks2017a}, to threshold out samples as out-of-distribution when the score is low. To define a score function for {{\method}} models, we first observe that the \emph{log-likelihood} of $\rvz_n$, which is only available for {{\method}} (and not for standard models), can be a strong score to detect novelties those are semantically far from in-distribution. Specifically, we use $\log \mathcal{N}(\rvz_n; 0, I) = -\tfrac{1}{2} \|\rvz_n \|^2$, as we assume that $\rvz$ follows isotropic Gaussian $\mathcal{N}(0, I)$. For detecting so-called ``harder'' novelties, we propose to use the log-likelihood score of $\rvy$ under a \emph{symmetric {Dirichlet} distribution} of parameter $\alpha > 0$, namely $\mathrm{Dir}_{\alpha}(\rvy) \in \Delta^{|\mathcal{Y}|-1}$, rather than simply using $\max_y p(y|x)$: \ie $\log \mathrm{Dir}_{\alpha}(\rvy) = (\alpha - 1) \sum_i \log y_i$. Note that the distribution gets closer to the {symmetric (discrete)} one-hot distribution as $\alpha \rightarrow 0$, {which makes sense for most classification tasks}, and here we simply use $\alpha=0.05$ throughout experiments.\footnote{{In practice, we observe that other choices in a moderate range of $\alpha$ near 0 do not much affect performance.}} 

We consider two evaluation benchmarks: (a) the \emph{``standard''} benchmark, that has been actively adopted in the literature \cite{hendrycks2017a,liang2018enhancing,lee2018maha}, assumes the CIFAR-10 test set as in-distribution and measures the detection performance of other independent datasets; (b) a recent \emph{OBJECTS} benchmark  \cite{yang2022fsood}, on the other hand, extends the CIFAR-10 benchmark to also consider ``near'' in-distribution in OOD evaluation. Specifically, OBJECTS assumes CIFAR-10-C \cite{hendrycks2018benchmarking} and ImageNet-10 as in-distribution in test-time as well as CIFAR-10, making the detection much more challenging. In this experiment, we compare ResNet-18 \cite{he2016deep} models trained on CIFAR-10 following the setup of \cite{yang2022fsood}.

The results are reported in Table~\ref{tab:ood} and \ref{tab:newood_far} for the standard and OBJECTS benchmarks, respectively. Overall, we confirm that the score function combining the information of $\rvz_n$ and $\rvy$ of {{\method}} significantly improves novelty detection in a complementary manner over strong baselines, showing the effectiveness of modeling nuisance. For example, in Table~\ref{tab:ood}, the combined score achieves near-perfect AUROCs for detecting SVHN, LSUN and ImageNet datasets. Regarding Table~\ref{tab:newood_far}, on the other hand,  {{\method}} improves the previous best AUROC (of Mahalanobis \cite{lee2018maha}) on OBJECTS \emph{vs.}~MNIST from $77.04 \rightarrow 92.43$. The improved results on OBJECTS imply that both of the representation and score obtained from {{\method}} help to better discriminate in- \emph{vs.}~out-of-distribution in more \emph{semantic} senses.

\subsection{Certified adversarial robustness}
\label{exp:adv}

We also evaluate adversarial robustness \cite{szegedy2014intriguing,goodfellow2014explaining,madry2018towards} adopting the \emph{randomized smoothing} framework \cite{lecuyer2019certified,cohen2019certified} that can measure a \emph{certified} robustness for a given representation. Specifically, any classifier can be robustified by averaging its predictions under Gaussian noise, where the robustness at input $x$ depends on how consistent the classifier is on classifying $\mathcal{N}(x, \sigma^2 \mathbf{I})$ \cite{jeong2020consistency}. {Under this evaluation protocol, we suggest that adversarially-robust representations can be a natural byproduct from {\method} when combined with the randomized smoothing technique, without using any thorough adversarial training methods \cite{madry2018towards} that often require  significant training cost with no certification on the robustness.}

We follow the standard certification protocol \cite{cohen2019certified} to compare the \emph{certified test accuracy at radius $r$}, which is defined by the fraction of the test samples that a smoothed classifier classifies correctly with its certified radius larger than $r$. {We consider ViT-S models on CIFAR-10, and assume $\sigma=0.1$ for this experiment. The results summarized in Figure~\ref{figure:adv_robustness} show that our proposed {{\method}} achieves significantly better certified robustness compared to the baselines at all radii tested: \eg it improves certified robust accuracy of VIB by $39.6\% \rightarrow 56.8\%$ at $\varepsilon=0.1$. Again, the robustness obtained from {\method} is not from specific knowledge on the threat model, which implies that {\method} could offer \emph{free} adversarial robustness when combined with randomized smoothing.} This confirms that the robustness of {{\method}} is not only significant but also consistent \emph{per input}, especially considering its high certified robustness at higher $r$'s.

\begin{figure}[t]
\begin{subfigure}{\linewidth}
    \centering
    \includegraphics[width=0.85\linewidth]{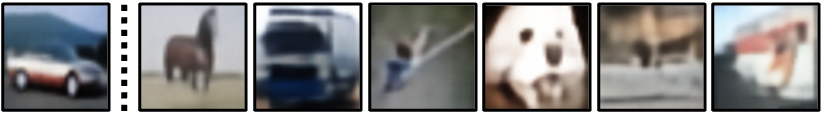}
    \caption{DisenIB \cite{pan2021disentangled}}\label{figure:swap_zn_disenib}
\end{subfigure}
\begin{subfigure}{\linewidth}
    \centering
    \includegraphics[width=0.85\linewidth]{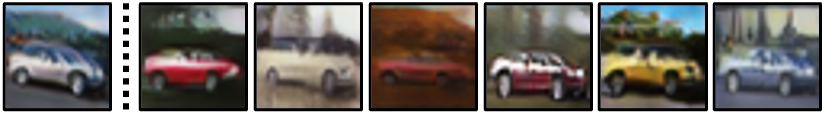}
    \caption{AENIB (Ours)}\label{figure:swap_zn_aenib}
\end{subfigure}
\vspace{-0.05in}
\caption{Comparison of CIFAR-10 reconstructions when the nuisance $\rvz_n$ is swapped with those of another (random) sample.}\label{figure:swap_zn}
\vspace{-0.1in}
\end{figure}

\subsection{{Comparison with DisenIB \cite{pan2021disentangled}}}
\label{exp:disenib}

In this section, we provide a comparison of {\method} with a related work of DisenIB \cite{pan2021disentangled}, as well as other variants of VIB, namely Nonlinear-VIB \cite{kolchinsky2019nonlinear} and Squared-VIB \cite{thobaben2020convex}. Here, DisenIB is a variant of IB which also considers a nuisance modeling (based on FactorVAE \cite{kim2018disentangling}), in a purpose of supervised disentangling. Specifically, DisenIB considers two independent encoders $\rvz := f(\rvx)$ and $\rvz_n := g(\rvx)$, and aims to optimize the following objective:
\begin{equation}\label{eq:disenib}
    \max_{f,g} I(\rvz;\rvy) + I(\rvx; \rvz_n, \rvy) - I(\rvz_n; \rvz). 
\end{equation}
Compared to our proposed NIB \eqref{eq:nib}, the most important difference between the two objectives is in their ``reconstruction'' terms: \ie $I(\rvx; \rvz_n, \rvy)$ of \eqref{eq:disenib} \textit{vs.}~$I(\rvx; \rvz, \rvz_n)$ of ours \eqref{eq:nib}. Due to this difference, the DisenIB objective \eqref{eq:disenib} cannot rule out the cases when $\rvz$ only encode few of ``shortcut'' signals in $\rvx$ (correlated to $\rvy$) even at optimum, in contrast to our key motivation of NIB \eqref{eq:nib} that aims to let $\rvz$ to encode \emph{every} $\rvy$-correlated signal in $\rvx$ as much as possible.

{We conduct experimental comparisons based on (a) our CIFAR-10 setups (Table~\ref{tab:corruption_vit}), and (b) directly upon the official implementation\footnote{\url{https://github.com/PanZiqiAI/disentangled-information-bottleneck}} of DisenIB on MNIST \cite{dataset/mnist} (Table~\ref{tab:mnistc} of Appendix~\ref{appendix:mnistc}). Here, we adopt and extend the benchmark to also cover \mbox{\emph{MNIST-C}} \cite{mu2019mnist}, a corrupted version of the MNIST. For the latter comparison, we use a simple 4-layer convolutional network as the encoder architecture. We train every MNIST model here for 100K updates and follow the other training details from the CIFAR experiments (see Appendix~\ref{appendix:setup:detail}).}

Overall, we observe that the effectiveness of {\method} still applies to these benchmarks. This is in contrast to DisenIB, given that the effectiveness from DisenIB, \eg its gain in AUROC on detecting Gaussian noise as an OOD (as conducted by \citet{pan2021disentangled}), could not be further generalized on CIFAR-10-C or MNIST-C, where {{\method}} still improves on as well as achieving the perfect score at the same OOD task. In Figure~\ref{figure:swap_zn_disenib}, we further observe qualitatively that DisenIB often leaves highly semantic information in the nuisance $\rvz_n$: its reconstruction can be completely changed by swapping $\rvz_n$ with those of another sample. This is essentially what AENIB addresses, as compared in Figure~\ref{figure:swap_zn_aenib}.

\section{Related work}

\vspace{0.05in}
\noindent\textbf{Out-of-distribution robustness. } 
Since the seminal works \cite{szegedy2014intriguing,Nguyen_2015_CVPR,amodei2016concrete} revealing the fragility of neural networks on out-of-distribution inputs, there have been significant attempts on identifying and improving various notions of robustness: \eg detecting novel inputs \cite{hendrycks2017a,lee2018maha,lee2018training,tack2020csi}, robustness against corruptions \cite{hendrycks2018benchmarking,geirhos2018imagenettrained,hendrycks2020augmix,xiao2021noise}, and adversarial noise \cite{madry2018towards,pmlr-v80-athalye18a,cohen2019certified,carlini2019evaluating}, to name a few. Due to its fundamental challenges in making neural network to extrapolate, however, most of the advances in the robustness literature has been made under assuming priors closely related to the individual problems: \eg an external data or data augmentations \cite{hendrycks2018deep,hendrycks2020augmix}, extra information from test-time samples \cite{wang2021tent}, or specific knowledge in threat models \cite{tramer2019adversarial,kang2019transfer}. In this work, we aim to improve multiple notions of robustness without assuming such priors, through a new training scheme that extends the standard information bottleneck principle under noisy observations in test-time. 

\vspace{0.05in}
\noindent\textbf{Hybrid generative-discriminative modeling. }
Our proposed method can be also viewed as a new approach of improving the robustness of discriminative models by incorporating a generative model, in the context that has been explored in recent works \cite{lee2018maha,schott2018towards,Grathwohl2020Your,Yang_2021_ICCV}. For example, \citet{lee2018maha, pmlr-v97-lee19f} have incorporated a simple (but of low expressivity for generation) Gaussian mixture model into discriminative classifiers; a line of research on \emph{Joint Energy-based Models} (JEM) \cite{Grathwohl2020Your,Yang_2021_ICCV} assumes an energy-based model but with a notable training instability for the purpose. In this work, we propose an autoencoder-based model to avoid such training instability, and consider a design that the \emph{nuisance} can succinctly supplement the given discriminative representation to be generative. We demonstrate that our approach can take the best of two worlds; it enables (a) stable training, while (b) attaining the high expressive generative performances.

\vspace{0.05in}
\noindent\textbf{Nuisance modeling. } The idea of incorporating nuisances can be also considered in the context of \emph{invertible} modeling, or as known as \emph{flow-based models} \cite{dinh2016density,kingma2018glow,behrmann2019invertible,grathwohl2019ffjord}, where the nuisance can be defined by splitting the (full-information) encoding $\rvz$ for a given subspace of interest as explored by \citet{jacobsen2018excessive,lynton2020training}. Unlike such approaches, our autoencoder-based nuisance modeling does not focus on the ``full'' invertibility for arbitrary inputs, but rather on inverting the data manifold given, which enabled (a) a much flexible encoder design in practice, and (b) a more scalable generative modeling of nuisance $\rvz_n$, \eg beyond an MNIST-scale as done by \citet{jacobsen2018excessive}. Other related works \cite{jaiswal2018unsupervised, jaiswal2019discovery, pan2021disentangled} do introduce an encoder for nuisance factors, but the notion of nuisance-ness has been focused in terms of the independence to $\rvz$ (for the purpose of feature disentangling), rather than to $\rvy$ as we focus in this work (for the purpose of robustness): \eg DisenIB \cite{pan2021disentangled} applies FactorVAE \cite{kim2018disentangling} between semantic and nuisance embeddings to force their independence. 
Yet, the literature has been also questioned on whether the idea can be scaled-up beyond, \eg MNIST, and our work does explore and establish a practical design with recent architectures and datasets addressing diverse modern security metrics. 

{We provide more extensive and detailed discussions on related works in Appendix~\ref{appendix:related}.}

\section{Conclusion}
\label{s:conclusion}

We suggest that having a good \emph{nuisance model} can be a tangible approach to induce a reliable representation. We develop a practical method of learning deep nuisance representation from data, and show its effectiveness to improve diverse reliability measures under a challenging setup of assuming no prior \cite{taori2020measuring}. We believe our work can be a useful step towards better understanding of out-of-distribution generalization in deep learning. Although the current scope is on a particular design of autoencoder based models, our framework of \emph{nuisance-extended IB} is not limited to it and future works could consider more diverse implementations. {Ultimately, we aim to approximate a challenging form of adversarial training with a mutual information constraint, which we believe will be a promising direction to explore.}

\section*{Acknowledgments}
This work was partly supported by Center for Applied Research in Artificial Intelligence (CARAI) grant funded by Defense Acquisition Program Administration (DAPA) and Agency for Defense Development (ADD) (UD190031RD), and by Institute of Information \& communications Technology Planning \& Evaluation (IITP) grant funded by the Korea government (MSIT) (No.2021-0-02068, Artificial Intelligence Innovation Hub; No.2019-0-00075, Artificial Intelligence Graduate School Program (KAIST)). We thank Subin Kim for the proofreading of our manuscript.

{\small
\bibliographystyle{ieee_fullname}
\bibliography{references}

\begin{thebibliography}{100}\itemsep=-1pt

\bibitem{alemi2016deep}
Alexander~A Alemi, Ian Fischer, Joshua~V Dillon, and Kevin Murphy.
\newblock Deep variational information bottleneck.
\newblock In {\em International Conference on Learning Representations}, 2017.

\bibitem{amodei2016concrete}
Dario Amodei, Chris Olah, Jacob Steinhardt, Paul Christiano, John Schulman, and
  Dan Man{\'e}.
\newblock Concrete problems in {AI} safety.
\newblock {\em arXiv preprint arXiv:1606.06565}, 2016.

\bibitem{aneja2021contrastive}
Jyoti Aneja, Alex Schwing, Jan Kautz, and Arash Vahdat.
\newblock A contrastive learning approach for training variational autoencoder
  priors.
\newblock {\em Advances in Neural Information Processing Systems}, 34, 2021.

\bibitem{lynton2020training}
Lynton Ardizzone, Radek Mackowiak, Carsten Rother, and Ullrich K\"{o}the.
\newblock Training normalizing flows with the information bottleneck for
  competitive generative classification.
\newblock In H. Larochelle, M. Ranzato, R. Hadsell, M.F. Balcan, and H. Lin,
  editors, {\em Advances in Neural Information Processing Systems}, volume~33,
  pages 7828--7840. Curran Associates, Inc., 2020.

\bibitem{pmlr-v80-athalye18a}
Anish Athalye, Nicholas Carlini, and David Wagner.
\newblock Obfuscated gradients give a false sense of security: Circumventing
  defenses to adversarial examples.
\newblock In Jennifer Dy and Andreas Krause, editors, {\em Proceedings of the
  35th International Conference on Machine Learning}, volume~80 of {\em
  Proceedings of Machine Learning Research}, pages 274--283. PMLR, 10--15 Jul
  2018.

\bibitem{behrmann2019invertible}
Jens Behrmann, Will Grathwohl, Ricky~TQ Chen, David Duvenaud, and
  J{\"o}rn-Henrik Jacobsen.
\newblock Invertible residual networks.
\newblock In {\em International Conference on Machine Learning}, pages
  573--582. PMLR, 2019.

\bibitem{bell1995information}
Anthony~J Bell and Terrence~J Sejnowski.
\newblock An information-maximization approach to blind separation and blind
  deconvolution.
\newblock {\em Neural computation}, 7(6):1129--1159, 1995.

\bibitem{beyer2022better}
Lucas Beyer, Xiaohua Zhai, and Alexander Kolesnikov.
\newblock Better plain {ViT} baselines for {ImageNet-1k}.
\newblock {\em arXiv preprint arXiv:2205.01580}, 2022.

\bibitem{brock2021high}
Andy Brock, Soham De, Samuel~L Smith, and Karen Simonyan.
\newblock High-performance large-scale image recognition without normalization.
\newblock In {\em International Conference on Machine Learning}, pages
  1059--1071. PMLR, 2021.

\bibitem{brunet2011mathematical}
Dominique Brunet, Edward~R Vrscay, and Zhou Wang.
\newblock On the mathematical properties of the structural similarity index.
\newblock {\em IEEE Transactions on Image Processing}, 21(4):1488--1499, 2011.

\bibitem{carlini2019evaluating}
Nicholas Carlini, Anish Athalye, Nicolas Papernot, Wieland Brendel, Jonas
  Rauber, Dimitris Tsipras, Ian Goodfellow, Aleksander Madry, and Alexey
  Kurakin.
\newblock On evaluating adversarial robustness, 2019.

\bibitem{chalk2016relevant}
Matthew Chalk, Olivier Marre, and Gasper Tkacik.
\newblock Relevant sparse codes with variational information bottleneck.
\newblock In D. Lee, M. Sugiyama, U. Luxburg, I. Guyon, and R. Garnett,
  editors, {\em Advances in Neural Information Processing Systems}, volume~29.
  Curran Associates, Inc., 2016.

\bibitem{chen2019residual}
Ricky~TQ Chen, Jens Behrmann, David~K Duvenaud, and J{\"o}rn-Henrik Jacobsen.
\newblock Residual flows for invertible generative modeling.
\newblock {\em Advances in Neural Information Processing Systems}, 32, 2019.

\bibitem{chen2020simclr}
Ting Chen, Simon Kornblith, Mohammad Norouzi, and Geoffrey Hinton.
\newblock A simple framework for contrastive learning of visual
  representations.
\newblock In {\em Proceedings of the 37th International Conference on Machine
  Learning}, Proceedings of Machine Learning Research. PMLR, 2020.

\bibitem{child2021very}
Rewon Child.
\newblock Very deep {VAEs} generalize autoregressive models and can outperform
  them on images.
\newblock In {\em International Conference on Learning Representations}, 2021.

\bibitem{chun2020empirical}
Sanghyuk Chun, Seong~Joon Oh, Sangdoo Yun, Dongyoon Han, Junsuk Choe, and
  Youngjoon Yoo.
\newblock An empirical evaluation on robustness and uncertainty of
  regularization methods.
\newblock {\em arXiv preprint arXiv:2003.03879}, 2020.

\bibitem{cohen2019certified}
Jeremy Cohen, Elan Rosenfeld, and Zico Kolter.
\newblock Certified adversarial robustness via randomized smoothing.
\newblock In {\em International Conference on Machine Learning}, pages
  1310--1320. PMLR, 2019.

\bibitem{cubuk2020randaug}
Ekin~Dogus Cubuk, Barret Zoph, Jon Shlens, and Quoc Le.
\newblock Rand{Augment}: Practical automated data augmentation with a reduced
  search space.
\newblock In {\em Advances in Neural Information Processing Systems},
  volume~33, pages 18613--18624. Curran Associates, Inc., 2020.

\bibitem{dai2018diagnosing}
Bin Dai and David Wipf.
\newblock Diagnosing and enhancing {VAE} models.
\newblock In {\em International Conference on Learning Representations}, 2019.

\bibitem{dai2021coatnet}
Zihang Dai, Hanxiao Liu, Quoc~V Le, and Mingxing Tan.
\newblock Coatnet: Marrying convolution and attention for all data sizes.
\newblock {\em Advances in Neural Information Processing Systems},
  34:3965--3977, 2021.

\bibitem{darlow2018cinic}
Luke~N Darlow, Elliot~J Crowley, Antreas Antoniou, and Amos~J Storkey.
\newblock {CINIC-10} is not {ImageNet} or {CIFAR-10}.
\newblock {\em arXiv preprint arXiv:1810.03505}, 2018.

\bibitem{diffenderfer2021winning}
James Diffenderfer, Brian Bartoldson, Shreya Chaganti, Jize Zhang, and Bhavya
  Kailkhura.
\newblock A winning hand: Compressing deep networks can improve
  out-of-distribution robustness.
\newblock {\em Advances in Neural Information Processing Systems}, 34, 2021.

\bibitem{dinh2016density}
Laurent Dinh, Jascha Sohl-Dickstein, and Samy Bengio.
\newblock Density estimation using {Real NVP}, 2016.

\bibitem{dosovitskiy2021an}
Alexey Dosovitskiy, Lucas Beyer, Alexander Kolesnikov, Dirk Weissenborn,
  Xiaohua Zhai, Thomas Unterthiner, Mostafa Dehghani, Matthias Minderer, Georg
  Heigold, Sylvain Gelly, Jakob Uszkoreit, and Neil Houlsby.
\newblock An image is worth 16x16 words: Transformers for image recognition at
  scale.
\newblock In {\em International Conference on Learning Representations}, 2021.

\bibitem{Fetaya2020Understanding}
Ethan Fetaya, Joern-Henrik Jacobsen, Will Grathwohl, and Richard Zemel.
\newblock Understanding the limitations of conditional generative models.
\newblock In {\em International Conference on Learning Representations}, 2020.

\bibitem{geirhos2020shortcut}
Robert Geirhos, J{\"o}rn-Henrik Jacobsen, Claudio Michaelis, Richard Zemel,
  Wieland Brendel, Matthias Bethge, and Felix~A Wichmann.
\newblock Shortcut learning in deep neural networks.
\newblock {\em Nature Machine Intelligence}, 2(11):665--673, 2020.

\bibitem{geirhos2018imagenettrained}
Robert Geirhos, Patricia Rubisch, Claudio Michaelis, Matthias Bethge, Felix~A.
  Wichmann, and Wieland Brendel.
\newblock Imagenet-trained {CNN}s are biased towards texture; increasing shape
  bias improves accuracy and robustness.
\newblock In {\em International Conference on Learning Representations}, 2019.

\bibitem{goodfellow2014gan}
Ian Goodfellow, Jean Pouget-Abadie, Mehdi Mirza, Bing Xu, David Warde-Farley,
  Sherjil Ozair, Aaron Courville, and Yoshua Bengio.
\newblock Generative adversarial nets.
\newblock In Z. Ghahramani, M. Welling, C. Cortes, N.~D. Lawrence, and K.~Q.
  Weinberger, editors, {\em Advances in Neural Information Processing Systems
  27}, pages 2672--2680. Curran Associates, Inc., 2014.

\bibitem{goodfellow2014explaining}
Ian~J Goodfellow, Jonathon Shlens, and Christian Szegedy.
\newblock Explaining and harnessing adversarial examples.
\newblock In {\em International Conference on Learning Representations}, 2015.

\bibitem{grathwohl2019ffjord}
Will Grathwohl, Ricky T.~Q. Chen, Jesse Bettencourt, Ilya Sutskever, and David
  Duvenaud.
\newblock {FFJORD}: Free-form continuous dynamics for scalable reversible
  generative models.
\newblock {\em International Conference on Learning Representations}, 2019.

\bibitem{Grathwohl2020Your}
Will Grathwohl, Kuan-Chieh Wang, Joern-Henrik Jacobsen, David Duvenaud,
  Mohammad Norouzi, and Kevin Swersky.
\newblock Your classifier is secretly an energy based model and you should
  treat it like one.
\newblock In {\em International Conference on Learning Representations}, 2020.

\bibitem{gu2022vector}
Shuyang Gu, Dong Chen, Jianmin Bao, Fang Wen, Bo Zhang, Dongdong Chen, Lu Yuan,
  and Baining Guo.
\newblock Vector quantized diffusion model for text-to-image synthesis.
\newblock In {\em Proceedings of the IEEE/CVF Conference on Computer Vision and
  Pattern Recognition}, pages 10696--10706, 2022.

\bibitem{he2016deep}
Kaiming He, Xiangyu Zhang, Shaoqing Ren, and Jian Sun.
\newblock Deep residual learning for image recognition.
\newblock In {\em Proceedings of the IEEE conference on computer vision and
  pattern recognition}, pages 770--778, 2016.

\bibitem{Hendrycks_2021_ICCV}
Dan Hendrycks, Steven Basart, Norman Mu, Saurav Kadavath, Frank Wang, Evan
  Dorundo, Rahul Desai, Tyler Zhu, Samyak Parajuli, Mike Guo, Dawn Song, Jacob
  Steinhardt, and Justin Gilmer.
\newblock The many faces of robustness: A critical analysis of
  out-of-distribution generalization.
\newblock In {\em Proceedings of the IEEE/CVF International Conference on
  Computer Vision}, pages 8340--8349, October 2021.

\bibitem{hendrycks2018benchmarking}
Dan Hendrycks and Thomas Dietterich.
\newblock Benchmarking neural network robustness to common corruptions and
  perturbations.
\newblock In {\em International Conference on Learning Representations}, 2019.

\bibitem{hendrycks2017a}
Dan Hendrycks and Kevin Gimpel.
\newblock A baseline for detecting misclassified and out-of-distribution
  examples in neural networks.
\newblock In {\em International Conference on Learning Representations}, 2017.

\bibitem{hendrycks2018deep}
Dan Hendrycks, Mantas Mazeika, and Thomas Dietterich.
\newblock Deep anomaly detection with outlier exposure.
\newblock In {\em International Conference on Learning Representations}, 2019.

\bibitem{hendrycks2019using}
Dan Hendrycks, Mantas Mazeika, Saurav Kadavath, and Dawn Song.
\newblock Using self-supervised learning can improve model robustness and
  uncertainty.
\newblock {\em Advances in Neural Information Processing Systems}, 32, 2019.

\bibitem{hendrycks2020augmix}
Dan Hendrycks, Norman Mu, Ekin~Dogus Cubuk, Barret Zoph, Justin Gilmer, and
  Balaji Lakshminarayanan.
\newblock Aug{Mix}: A simple method to improve robustness and uncertainty under
  data shift.
\newblock In {\em International Conference on Learning Representations}, 2020.

\bibitem{hendrycks2022pixmix}
Dan Hendrycks, Andy Zou, Mantas Mazeika, Leonard Tang, Bo Li, Dawn Song, and
  Jacob Steinhardt.
\newblock Pix{Mix}: Dreamlike pictures comprehensively improve safety measures.
\newblock In {\em Proceedings of the IEEE/CVF Conference on Computer Vision and
  Pattern Recognition}, pages 16783--16792, 2022.

\bibitem{higgins2016beta}
Irina Higgins, Loic Matthey, Arka Pal, Christopher Burgess, Xavier Glorot,
  Matthew Botvinick, Shakir Mohamed, and Alexander Lerchner.
\newblock beta-vae: Learning basic visual concepts with a constrained
  variational framework.
\newblock In {\em International Conference on Learning Representations}, 2016.

\bibitem{ho2022imagen}
Jonathan Ho, William Chan, Chitwan Saharia, Jay Whang, Ruiqi Gao, Alexey
  Gritsenko, Diederik~P Kingma, Ben Poole, Mohammad Norouzi, David~J Fleet,
  et~al.
\newblock Imagen video: High definition video generation with diffusion models.
\newblock {\em arXiv preprint arXiv:2210.02303}, 2022.

\bibitem{ho2020ddpm}
Jonathan Ho, Ajay Jain, and Pieter Abbeel.
\newblock Denoising diffusion probabilistic models.
\newblock In H. Larochelle, M. Ranzato, R. Hadsell, M.~F. Balcan, and H. Lin,
  editors, {\em Advances in Neural Information Processing Systems}, volume~33,
  pages 6840--6851. Curran Associates, Inc., 2020.

\bibitem{ilyas2019adversarial}
Andrew Ilyas, Shibani Santurkar, Dimitris Tsipras, Logan Engstrom, Brandon
  Tran, and Aleksander Madry.
\newblock Adversarial examples are not bugs, they are features.
\newblock In {\em Advances in Neural Information Processing Systems},
  volume~32. Curran Associates, Inc., 2019.

\bibitem{ioffe2015batch}
Sergey Ioffe and Christian Szegedy.
\newblock Batch normalization: Accelerating deep network training by reducing
  internal covariate shift.
\newblock In {\em International conference on machine learning}, pages
  448--456. PMLR, 2015.

\bibitem{jacobsen2018excessive}
Joern-Henrik Jacobsen, Jens Behrmann, Richard Zemel, and Matthias Bethge.
\newblock Excessive invariance causes adversarial vulnerability.
\newblock In {\em International Conference on Learning Representations}, 2019.

\bibitem{jacobsen2018irevnet}
Jörn-Henrik Jacobsen, Arnold~W.M. Smeulders, and Edouard Oyallon.
\newblock {i-RevNet}: Deep invertible networks.
\newblock In {\em International Conference on Learning Representations}, 2018.

\bibitem{jaiswal2019discovery}
Ayush Jaiswal, Rob Brekelmans, Daniel Moyer, Greg~Ver Steeg, Wael AbdAlmageed,
  and Premkumar Natarajan.
\newblock Discovery and separation of features for invariant representation
  learning, 2019.

\bibitem{jaiswal2018unsupervised}
Ayush Jaiswal, Rex~Yue Wu, Wael Abd-Almageed, and Prem Natarajan.
\newblock Unsupervised adversarial invariance.
\newblock {\em Advances in Neural Information Processing Systems}, 31, 2018.

\bibitem{jeong2020consistency}
Jongheon Jeong and Jinwoo Shin.
\newblock Consistency regularization for certified robustness of smoothed
  classifiers.
\newblock {\em Advances in Neural Information Processing Systems},
  33:10558--10570, 2020.

\bibitem{jeong2021training}
Jongheon Jeong and Jinwoo Shin.
\newblock Training {GAN}s with stronger augmentations via contrastive
  discriminator.
\newblock In {\em International Conference on Learning Representations}, 2021.

\bibitem{kang2019transfer}
Daniel Kang, Yi Sun, Tom Brown, Dan Hendrycks, and Jacob Steinhardt.
\newblock Transfer of adversarial robustness between perturbation types, 2019.

\bibitem{karras2020training}
Tero Karras, Miika Aittala, Janne Hellsten, Samuli Laine, Jaakko Lehtinen, and
  Timo Aila.
\newblock Training generative adversarial networks with limited data.
\newblock In H. Larochelle, M. Ranzato, R. Hadsell, M.F. Balcan, and H. Lin,
  editors, {\em Advances in Neural Information Processing Systems}, volume~33,
  pages 12104--12114. Curran Associates, Inc., 2020.

\bibitem{karras2019style}
Tero Karras, Samuli Laine, and Timo Aila.
\newblock A style-based generator architecture for generative adversarial
  networks.
\newblock In {\em Proceedings of the IEEE/CVF conference on computer vision and
  pattern recognition}, pages 4401--4410, 2019.

\bibitem{karras2020analyzing}
Tero Karras, Samuli Laine, Miika Aittala, Janne Hellsten, Jaakko Lehtinen, and
  Timo Aila.
\newblock Analyzing and improving the image quality of {StyleGAN}.
\newblock In {\em Proceedings of the IEEE/CVF Conference on Computer Vision and
  Pattern Recognition}, pages 8110--8119, 2020.

\bibitem{khosla2020supervised}
Prannay Khosla, Piotr Teterwak, Chen Wang, Aaron Sarna, Yonglong Tian, Phillip
  Isola, Aaron Maschinot, Ce Liu, and Dilip Krishnan.
\newblock Supervised contrastive learning.
\newblock {\em Advances in Neural Information Processing Systems},
  33:18661--18673, 2020.

\bibitem{kim2018disentangling}
Hyunjik Kim and Andriy Mnih.
\newblock Disentangling by factorising.
\newblock In {\em International Conference on Machine Learning}, pages
  2649--2658. PMLR, 2018.

\bibitem{kingma15adam}
Diederik~P. Kingma and Jimmy Ba.
\newblock Adam: A method for stochastic optimization.
\newblock In {\em ICLR (Poster)}, 2015.

\bibitem{kingma2018glow}
Durk~P Kingma and Prafulla Dhariwal.
\newblock Glow: Generative flow with invertible 1x1 convolutions.
\newblock In S. Bengio, H. Wallach, H. Larochelle, K. Grauman, N. Cesa-Bianchi,
  and R. Garnett, editors, {\em Advances in Neural Information Processing
  Systems}, volume~31. Curran Associates, Inc., 2018.

\bibitem{kingma2016improved}
Durk~P Kingma, Tim Salimans, Rafal Jozefowicz, Xi Chen, Ilya Sutskever, and Max
  Welling.
\newblock Improved variational inference with inverse autoregressive flow.
\newblock {\em Advances in neural information processing systems}, 29, 2016.

\bibitem{kingma2014autoencoding}
Diederik~P Kingma and Max Welling.
\newblock Auto-encoding variational bayes, 2014.

\bibitem{kireev2022effectiveness}
Klim Kireev, Maksym Andriushchenko, and Nicolas Flammarion.
\newblock On the effectiveness of adversarial training against common
  corruptions.
\newblock In {\em Uncertainty in Artificial Intelligence}, pages 1012--1021.
  PMLR, 2022.

\bibitem{kobyzev2020normalizing}
Ivan Kobyzev, Simon~JD Prince, and Marcus~A Brubaker.
\newblock Normalizing flows: An introduction and review of current methods.
\newblock {\em IEEE transactions on pattern analysis and machine intelligence},
  43(11):3964--3979, 2020.

\bibitem{kohonen1990self}
Teuvo Kohonen.
\newblock The self-organizing map.
\newblock {\em Proceedings of the IEEE}, 78(9):1464--1480, 1990.

\bibitem{kolchinsky2019nonlinear}
Artemy Kolchinsky, Brendan~D Tracey, and David~H Wolpert.
\newblock Nonlinear information bottleneck.
\newblock {\em Entropy}, 21(12):1181, 2019.

\bibitem{dataset/cifar}
Alex Krizhevsky.
\newblock Learning multiple layers of features from tiny images.
\newblock Technical report, Department of Computer Science, University of
  Toronto, 2009.

\bibitem{kriz2012alexnet}
Alex Krizhevsky, Ilya Sutskever, and Geoffrey~E Hinton.
\newblock Imagenet classification with deep convolutional neural networks.
\newblock In F. Pereira, C.J. Burges, L. Bottou, and K.Q. Weinberger, editors,
  {\em Advances in Neural Information Processing Systems}, volume~25. Curran
  Associates, Inc., 2012.

\bibitem{kurach2019large}
Karol Kurach, Mario Lucic, Xiaohua Zhai, Marcin Michalski, and Sylvain Gelly.
\newblock A large-scale study on regularization and normalization in {GAN}s.
\newblock In {\em International Conference on Machine Learning}, pages
  3581--3590. PMLR, 2019.

\bibitem{dataset/mnist}
Y. {Le{C}un}, L. {Bottou}, Y. {Bengio}, and P. {Haffner}.
\newblock Gradient-based learning applied to document recognition.
\newblock {\em Proceedings of the IEEE}, 86(11):2278--2324, Nov 1998.

\bibitem{lecuyer2019certified}
Mathias Lecuyer, Vaggelis Atlidakis, Roxana Geambasu, Daniel Hsu, and Suman
  Jana.
\newblock Certified robustness to adversarial examples with differential
  privacy.
\newblock In {\em 2019 IEEE Symposium on Security and Privacy (SP)}, pages
  656--672. IEEE, 2019.

\bibitem{lee2018training}
Kimin Lee, Honglak Lee, Kibok Lee, and Jinwoo Shin.
\newblock Training confidence-calibrated classifiers for detecting
  out-of-distribution samples.
\newblock In {\em International Conference on Learning Representations}, 2018.

\bibitem{lee2018maha}
Kimin Lee, Kibok Lee, Honglak Lee, and Jinwoo Shin.
\newblock A simple unified framework for detecting out-of-distribution samples
  and adversarial attacks.
\newblock In S. Bengio, H. Wallach, H. Larochelle, K. Grauman, N. Cesa-Bianchi,
  and R. Garnett, editors, {\em Advances in Neural Information Processing
  Systems}, volume~31. Curran Associates, Inc., 2018.

\bibitem{pmlr-v97-lee19f}
Kimin Lee, Sukmin Yun, Kibok Lee, Honglak Lee, Bo Li, and Jinwoo Shin.
\newblock Robust inference via generative classifiers for handling noisy
  labels.
\newblock In Kamalika Chaudhuri and Ruslan Salakhutdinov, editors, {\em
  Proceedings of the 36th International Conference on Machine Learning},
  volume~97 of {\em Proceedings of Machine Learning Research}, pages
  3763--3772. PMLR, 09--15 Jun 2019.

\bibitem{liang2018enhancing}
Shiyu Liang, Yixuan Li, and R. Srikant.
\newblock Enhancing the reliability of out-of-distribution image detection in
  neural networks.
\newblock In {\em International Conference on Learning Representations}, 2018.

\bibitem{liu2021towards}
Bingchen Liu, Yizhe Zhu, Kunpeng Song, and Ahmed Elgammal.
\newblock Towards faster and stabilized {GAN} training for high-fidelity
  few-shot image synthesis.
\newblock In {\em International Conference on Learning Representations}, 2021.

\bibitem{liu2020energy}
Weitang Liu, Xiaoyun Wang, John Owens, and Yixuan Li.
\newblock Energy-based out-of-distribution detection.
\newblock {\em Advances in Neural Information Processing Systems},
  33:21464--21475, 2020.

\bibitem{liu2015faceattributes}
Ziwei Liu, Ping Luo, Xiaogang Wang, and Xiaoou Tang.
\newblock Deep learning face attributes in the wild.
\newblock In {\em Proceedings of International Conference on Computer Vision},
  December 2015.

\bibitem{loshchilov2016sgdr}
Ilya Loshchilov and Frank Hutter.
\newblock {SGDR}: Stochastic gradient descent with warm restarts, 2016.

\bibitem{loshchilov2018decoupled}
Ilya Loshchilov and Frank Hutter.
\newblock Decoupled weight decay regularization.
\newblock In {\em International Conference on Learning Representations}, 2019.

\bibitem{louizos2015variational}
Christos Louizos, Kevin Swersky, Yujia Li, Max Welling, and Richard Zemel.
\newblock The variational fair autoencoder, 2015.

\bibitem{lu2020harder}
Shangyun Lu, Bradley Nott, Aaron Olson, Alberto Todeschini, Hossein Vahabi,
  Yair Carmon, and Ludwig Schmidt.
\newblock Harder or different? a closer look at distribution shift in dataset
  reproduction.
\newblock In {\em ICML Workshop on Uncertainty and Robustness in Deep
  Learning}, 2020.

\bibitem{madry2018towards}
Aleksander Madry, Aleksandar Makelov, Ludwig Schmidt, Dimitris Tsipras, and
  Adrian Vladu.
\newblock Towards deep learning models resistant to adversarial attacks.
\newblock In {\em International Conference on Learning Representations}, 2018.

\bibitem{makhzani2015adversarial}
Alireza Makhzani, Jonathon Shlens, Navdeep Jaitly, Ian Goodfellow, and Brendan
  Frey.
\newblock Adversarial autoencoders.
\newblock {\em arXiv preprint arXiv:1511.05644}, 2015.

\bibitem{mescheder2018training}
Lars Mescheder, Andreas Geiger, and Sebastian Nowozin.
\newblock Which training methods for {GANs} do actually converge?
\newblock In {\em International conference on machine learning}, pages
  3481--3490. PMLR, 2018.

\bibitem{mildenhall2022dark}
Ben Mildenhall, Peter Hedman, Ricardo Martin-Brualla, Pratul~P Srinivasan, and
  Jonathan~T Barron.
\newblock Nerf in the dark: High dynamic range view synthesis from noisy raw
  images.
\newblock In {\em Proceedings of the IEEE/CVF Conference on Computer Vision and
  Pattern Recognition}, pages 16190--16199, 2022.

\bibitem{mildenhall2021nerf}
Ben Mildenhall, Pratul~P Srinivasan, Matthew Tancik, Jonathan~T Barron, Ravi
  Ramamoorthi, and Ren Ng.
\newblock Nerf: Representing scenes as neural radiance fields for view
  synthesis.
\newblock {\em Communications of the ACM}, 65(1):99--106, 2021.

\bibitem{miller2021accuracy}
John~P Miller, Rohan Taori, Aditi Raghunathan, Shiori Sagawa, Pang~Wei Koh,
  Vaishaal Shankar, Percy Liang, Yair Carmon, and Ludwig Schmidt.
\newblock Accuracy on the line: on the strong correlation between
  out-of-distribution and in-distribution generalization.
\newblock In {\em International Conference on Machine Learning}, pages
  7721--7735. PMLR, 2021.

\bibitem{mu2019mnist}
Norman Mu and Justin Gilmer.
\newblock {MNIST-C}: A robustness benchmark for computer vision.
\newblock {\em arXiv preprint arXiv:1906.02337}, 2019.

\bibitem{nalisnick2018do}
Eric Nalisnick, Akihiro Matsukawa, Yee~Whye Teh, Dilan Gorur, and Balaji
  Lakshminarayanan.
\newblock Do deep generative models know what they don't know?
\newblock In {\em International Conference on Learning Representations}, 2019.

\bibitem{Nguyen_2015_CVPR}
Anh Nguyen, Jason Yosinski, and Jeff Clune.
\newblock Deep neural networks are easily fooled: High confidence predictions
  for unrecognizable images.
\newblock In {\em Proceedings of the IEEE Conference on Computer Vision and
  Pattern Recognition (CVPR)}, June 2015.

\bibitem{oord2018representation}
Aaron van~den Oord, Yazhe Li, and Oriol Vinyals.
\newblock Representation learning with contrastive predictive coding.
\newblock {\em arXiv preprint arXiv:1807.03748}, 2018.

\bibitem{pan2021disentangled}
Ziqi Pan, Li Niu, Jianfu Zhang, and Liqing Zhang.
\newblock Disentangled information bottleneck.
\newblock In {\em Proceedings of the AAAI Conference on Artificial
  Intelligence}, volume~35, pages 9285--9293, 2021.

\bibitem{parmar2021dual}
Gaurav Parmar, Dacheng Li, Kwonjoon Lee, and Zhuowen Tu.
\newblock Dual contradistinctive generative autoencoder.
\newblock In {\em Proceedings of the IEEE/CVF Conference on Computer Vision and
  Pattern Recognition}, pages 823--832, 2021.

\bibitem{recht2018cifar10.1}
Benjamin Recht, Rebecca Roelofs, Ludwig Schmidt, and Vaishaal Shankar.
\newblock Do {CIFAR-10} classifiers generalize to {CIFAR-10}?
\newblock {\em arXiv preprint arXiv:1806.00451}, 2018.

\bibitem{ren2019lratio}
Jie Ren, Peter~J. Liu, Emily Fertig, Jasper Snoek, Ryan Poplin, Mark Depristo,
  Joshua Dillon, and Balaji Lakshminarayanan.
\newblock Likelihood ratios for out-of-distribution detection.
\newblock In H. Wallach, H. Larochelle, A. Beygelzimer, F. d\textquotesingle
  Alch\'{e}-Buc, E. Fox, and R. Garnett, editors, {\em Advances in Neural
  Information Processing Systems}, volume~32. Curran Associates, Inc., 2019.

\bibitem{rombach2022high}
Robin Rombach, Andreas Blattmann, Dominik Lorenz, Patrick Esser, and Bj{\"o}rn
  Ommer.
\newblock High-resolution image synthesis with latent diffusion models.
\newblock In {\em Proceedings of the IEEE/CVF Conference on Computer Vision and
  Pattern Recognition}, pages 10684--10695, 2022.

\bibitem{dataset/ilsvrc}
Olga Russakovsky, Jia Deng, Hao Su, Jonathan Krause, Sanjeev Satheesh, Sean Ma,
  Zhiheng Huang, Andrej Karpathy, Aditya Khosla, Michael Bernstein,
  Alexander~C. Berg, and Li Fei-Fei.
\newblock {ImageNet Large Scale Visual Recognition Challenge}.
\newblock {\em International Journal of Computer Vision}, 115(3):211--252,
  2015.

\bibitem{sauer2021projected}
Axel Sauer, Kashyap Chitta, Jens M{\"u}ller, and Andreas Geiger.
\newblock Projected {GAN}s converge faster.
\newblock In A. Beygelzimer, Y. Dauphin, P. Liang, and J.~Wortman Vaughan,
  editors, {\em Advances in Neural Information Processing Systems}, 2021.

\bibitem{schott2018towards}
Lukas Schott, Jonas Rauber, Matthias Bethge, and Wieland Brendel.
\newblock Towards the first adversarially robust neural network model on
  {MNIST}.
\newblock In {\em International Conference on Learning Representations}, 2019.

\bibitem{Serra2020Input}
Joan Serra, David Alvarez, Vicenc Gomez, Olga Slizovskaia, Jose~F. Nunez, and
  Jordi Luque.
\newblock Input complexity and out-of-distribution detection with
  likelihood-based generative models.
\newblock In {\em International Conference on Learning Representations}, 2020.

\bibitem{song2021denoising}
Jiaming Song, Chenlin Meng, and Stefano Ermon.
\newblock Denoising diffusion implicit models.
\newblock In {\em International Conference on Learning Representations}, 2021.

\bibitem{szegedy2014intriguing}
Christian Szegedy, Wojciech Zaremba, Ilya Sutskever, Joan Bruna, Dumitru Erhan,
  Ian Goodfellow, and Rob Fergus.
\newblock Intriguing properties of neural networks, 2014.

\bibitem{tack2020csi}
Jihoon Tack, Sangwoo Mo, Jongheon Jeong, and Jinwoo Shin.
\newblock {CSI}: Novelty detection via contrastive learning on distributionally
  shifted instances.
\newblock In {\em Advances in Neural Information Processing Systems}, 2020.

\bibitem{tancik2022block}
Matthew Tancik, Vincent Casser, Xinchen Yan, Sabeek Pradhan, Ben Mildenhall,
  Pratul~P Srinivasan, Jonathan~T Barron, and Henrik Kretzschmar.
\newblock Block-nerf: Scalable large scene neural view synthesis.
\newblock In {\em Proceedings of the IEEE/CVF Conference on Computer Vision and
  Pattern Recognition}, pages 8248--8258, 2022.

\bibitem{taori2020measuring}
Rohan Taori, Achal Dave, Vaishaal Shankar, Nicholas Carlini, Benjamin Recht,
  and Ludwig Schmidt.
\newblock Measuring robustness to natural distribution shifts in image
  classification.
\newblock {\em Advances in Neural Information Processing Systems},
  33:18583--18599, 2020.

\bibitem{thobaben2020convex}
R Thobaben, M Skoglund, et~al.
\newblock The convex information bottleneck lagrangian.
\newblock {\em Entropy}, 22(1), 2020.

\bibitem{tishby99information}
Naftali Tishby, Fernando~C. Pereira, and William Bialek.
\newblock The information bottleneck method.
\newblock In {\em Proceedings of the 37-th Annual Allerton Conference on
  Communication, Control and Computing}, pages 368--377, 1999.

\bibitem{naf}
Naftali Tishby and Noga Zaslavsky.
\newblock Deep learning and the information bottleneck principle.
\newblock In {\em 2015 IEEE Information Theory Workshop}, pages 1--5, 2015.

\bibitem{tong2022videomae}
Zhan Tong, Yibing Song, Jue Wang, and Limin Wang.
\newblock Videomae: Masked autoencoders are data-efficient learners for
  self-supervised video pre-training.
\newblock {\em arXiv preprint arXiv:2203.12602}, 2022.

\bibitem{torralba2008tiny}
Antonio Torralba, Rob Fergus, and William~T. Freeman.
\newblock 80 million tiny images: A large data set for nonparametric object and
  scene recognition.
\newblock {\em IEEE Transactions on Pattern Analysis and Machine Intelligence},
  30(11):1958--1970, 2008.

\bibitem{touvron2021training}
Hugo Touvron, Matthieu Cord, Matthijs Douze, Francisco Massa, Alexandre
  Sablayrolles, and Herv{\'e} J{\'e}gou.
\newblock Training data-efficient image transformers \& distillation through
  attention.
\newblock In {\em International Conference on Machine Learning}, pages
  10347--10357. PMLR, 2021.

\bibitem{tramer2019adversarial}
Florian Tramer and Dan Boneh.
\newblock Adversarial training and robustness for multiple perturbations.
\newblock {\em Advances in Neural Information Processing Systems}, 32, 2019.

\bibitem{vahdat2020nvae}
Arash Vahdat and Jan Kautz.
\newblock {NVAE}: A deep hierarchical variational autoencoder.
\newblock {\em Advances in Neural Information Processing Systems},
  33:19667--19679, 2020.

\bibitem{oord2017neural}
Aaron van~den Oord, Oriol Vinyals, and koray kavukcuoglu.
\newblock Neural discrete representation learning.
\newblock In I. Guyon, U.~Von Luxburg, S. Bengio, H. Wallach, R. Fergus, S.
  Vishwanathan, and R. Garnett, editors, {\em Advances in Neural Information
  Processing Systems}, volume~30. Curran Associates, Inc., 2017.

\bibitem{pascal2008denoising}
Pascal Vincent, Hugo Larochelle, Yoshua Bengio, and Pierre-Antoine Manzagol.
\newblock Extracting and composing robust features with denoising autoencoders.
\newblock In {\em Proceedings of the 25th International Conference on Machine
  Learning}, page 1096–1103, New York, NY, USA, 2008. Association for
  Computing Machinery.

\bibitem{wang2021tent}
Dequan Wang, Evan Shelhamer, Shaoteng Liu, Bruno Olshausen, and Trevor Darrell.
\newblock Tent: Fully test-time adaptation by entropy minimization.
\newblock In {\em International Conference on Learning Representations}, 2021.

\bibitem{wang2019learning}
Haohan Wang, Songwei Ge, Zachary Lipton, and Eric~P Xing.
\newblock Learning robust global representations by penalizing local predictive
  power.
\newblock In {\em Advances in Neural Information Processing Systems}, pages
  10506--10518, 2019.

\bibitem{wang2004image}
Zhou Wang, Alan~C Bovik, Hamid~R Sheikh, and Eero~P Simoncelli.
\newblock Image quality assessment: from error visibility to structural
  similarity.
\newblock {\em IEEE Transactions on Image Processing}, 13(4):600--612, 2004.

\bibitem{xiao2021noise}
Kai~Yuanqing Xiao, Logan Engstrom, Andrew Ilyas, and Aleksander Madry.
\newblock Noise or signal: The role of image backgrounds in object recognition.
\newblock In {\em International Conference on Learning Representations}, 2021.

\bibitem{xiao2020lregret}
Zhisheng Xiao, Qing Yan, and Yali Amit.
\newblock Likelihood regret: An out-of-distribution detection score for
  variational auto-encoder.
\newblock In H. Larochelle, M. Ranzato, R. Hadsell, M.~F. Balcan, and H. Lin,
  editors, {\em Advances in Neural Information Processing Systems}, volume~33,
  pages 20685--20696. Curran Associates, Inc., 2020.

\bibitem{xie2020adversarial}
Cihang Xie, Mingxing Tan, Boqing Gong, Jiang Wang, Alan~L Yuille, and Quoc~V
  Le.
\newblock Adversarial examples improve image recognition.
\newblock In {\em Proceedings of the IEEE/CVF Conference on Computer Vision and
  Pattern Recognition}, pages 819--828, 2020.

\bibitem{yang2022fsood}
Jingkang Yang, Kaiyang Zhou, and Ziwei Liu.
\newblock Full-spectrum out-of-distribution detection.
\newblock {\em arXiv preprint arXiv:2204.05306}, 2022.

\bibitem{Yang_2021_ICCV}
Xiulong Yang and Shihao Ji.
\newblock {JEM}++: Improved techniques for training {JEM}.
\newblock In {\em Proceedings of the IEEE/CVF International Conference on
  Computer Vision}, pages 6494--6503, October 2021.

\bibitem{yu2022vectorquantized}
Jiahui Yu, Xin Li, Jing~Yu Koh, Han Zhang, Ruoming Pang, James Qin, Alexander
  Ku, Yuanzhong Xu, Jason Baldridge, and Yonghui Wu.
\newblock Vector-quantized image modeling with improved {VQGAN}.
\newblock In {\em International Conference on Learning Representations}, 2022.

\bibitem{yu2022generating}
Sihyun Yu, Jihoon Tack, Sangwoo Mo, Hyunsu Kim, Junho Kim, Jung-Woo Ha, and
  Jinwoo Shin.
\newblock Generating videos with dynamics-aware implicit generative adversarial
  networks.
\newblock In {\em International Conference on Learning Representations}, 2022.

\bibitem{yun2019cutmix}
Sangdoo Yun, Dongyoon Han, Seong~Joon Oh, Sanghyuk Chun, Junsuk Choe, and
  Youngjoon Yoo.
\newblock Cut{Mix}: Regularization strategy to train strong classifiers with
  localizable features.
\newblock In {\em Proceedings of the IEEE/CVF International Conference on
  Computer Vision}, pages 6023--6032, 2019.

\bibitem{zhang2018mixup}
Hongyi Zhang, Moustapha Cisse, Yann~N. Dauphin, and David Lopez-Paz.
\newblock {mixup}: Beyond empirical risk minimization.
\newblock In {\em International Conference on Learning Representations}, 2018.

\bibitem{zhang2019theoretically}
Hongyang Zhang, Yaodong Yu, Jiantao Jiao, Eric Xing, Laurent El~Ghaoui, and
  Michael Jordan.
\newblock Theoretically principled trade-off between robustness and accuracy.
\newblock In {\em International Conference on Machine Learning}, pages
  7472--7482. PMLR, 2019.

\bibitem{zhang2020perceptual}
Zijun Zhang, Ruixiang Zhang, Zongpeng Li, Yoshua Bengio, and Liam Paull.
\newblock Perceptual generative autoencoders.
\newblock In {\em International Conference on Machine Learning}, pages
  11298--11306. PMLR, 2020.

\bibitem{zhao2020diffaugment}
Shengyu Zhao, Zhijian Liu, Ji Lin, Jun-Yan Zhu, and Song Han.
\newblock Differentiable augmentation for data-efficient {GAN} training.
\newblock {\em Advances in Neural Information Processing Systems},
  33:7559--7570, 2020.

\end{thebibliography}
}

\clearpage
\appendix
\onecolumn

\begin{center}{\bf {\LARGE Supplementary Material}}
\end{center}
\begin{center}{\bf {\Large Enhancing Multiple Reliability Measures via \\ Nuisance-extended Information Bottleneck}}
\end{center}

\FloatBarrier
\section{Training procedure of {{\method}}}
\label{appendix:alg}
\begin{algorithm}[h]
\caption{Autoencoder-based nuisance-extended information bottleneck ({\method})}
\label{alg:training}
\begin{algorithmic}[1]
\REQUIRE encoder $f$, decoder $g$, discriminators $d$, prior $p_0(\rvz)$, $\alpha, \beta > 0$.
\vspace{0.05in}
\hrule
\vspace{0.05in}
\FOR{\# training iterations}
    \STATE Sample $(x_i, y_i)_{i=1}^{m} \sim p_d(\rvx, \rvy)$
    \STATE $\rvz^{(i)}, \rvz^{(i)}_n \leftarrow f(x_i)$, and sample $z^{(i)}, z^{(i)}_n \sim \rvz^{(i)}, \rvz^{(i)}_n$
    \STATE $\hat{x}_i \leftarrow g(z^{(i)}, z^{(i)}_n)$
    \STATE \textcolor{Blue9}{\textsc{// Update discriminators}}
    \STATE $L_{\tt ind} \leftarrow \mathbb{E}_{\rvz, \rvz_n \sim \mathcal{N}(0, I)} [\log d_\rvz(\rvz, \rvz_n)] + \frac{1}{m} \sum_{i} \log( 1 - d_{\rvz}(z, z_n))$
    \STATE $L_{\tt nuis}^D \leftarrow \frac{1}{m} \sum_{i} \mathbb{CE}(q_n(\rvy | z_n^{(i)}), y_i)$
    \STATE $L_D \leftarrow L_{\tt nuis}^D - L_{\tt ind}$
    \STATE $d_{\rvz}, q_n \leftarrow$ Update $d_{\rvz}, q_n$ to minimize $L_D$
    \STATE \textcolor{Blue9}{\textsc{// Update encoder and decoder}}
    \STATE $L_{\tt VIB}^{\beta} \leftarrow \frac{1}{m} \sum_{i} \left[-\log q(y_i|z_i) + \beta \mathrm{KL}(p(\rvz|x_i)\|p_0(\rvz))\right]$
    \STATE $L_{\tt recon} \leftarrow \frac{1}{m} \sum_{i} \frac{1}{\|x_i\|^2}\|x_i - \hat{x}_i\|_2^2$
    \STATE $L_{\tt nuis} \leftarrow \frac{1}{m} \sum_{i} \mathbb{CE}(q^*_n(\rvy | z_n^{(i)}), \frac{1}{|\mathcal{Y}|})$
    \STATE $L_{\tt {\method}} \leftarrow L_{\tt VIB}^{\beta} + \alpha L_{\tt recon} + L_{\tt nuis} + L_{\tt ind}$
    \STATE $f,g,\Pi_f \leftarrow$ Update $f, g, \Pi_f$ to minimize $L_{\tt {\method}}$
\ENDFOR
\end{algorithmic}
\end{algorithm}
\vspace{-0.05in}

\FloatBarrier
\section{Experimental details}
\label{appendix:setup}

\subsection{Training details}
\label{appendix:setup:detail}

Unless otherwise noted, we train each model for 200K updates for CIFAR-10 models, and 1M updates for ImageNet models. For training {{\method}} models, we use $\alpha=10.0, \beta=0.0001$ unless otherwise noted. We use different training configurations depending on the encoder architecture, \ie whether is it ResNet or ViT: (a) For ResNet-based models, we train the encoder part ($f$) via stochastic gradient descent (SGD) with batch size of 64 using Nesterov momentum of weight 0.9 without dampening. We set a weight decay of $10^{-4}$, and use the cosine learning rate scheduling \cite{loshchilov2016sgdr} from the initial learning rate of 0.1. For the remainder parts of our {{\method}} architecture, \eg the decoder $g$ and discriminator MLPs, on the other hand, we follow the training practices of GAN instead: specifically, we use Adam \cite{kingma15adam} with $(\alpha, \beta_1, \beta_2)=(0.0002, 0.5, 0.999)$, following the hyperparameter practices explored by \citet{kurach2019large}. (b) For ViT-based models, on the other hand, we train both (transformer-based) encoder and decoder models via AdamW \cite{loshchilov2018decoupled} with a weight decay of $10^{-4}$, using batch size 128 and $(\alpha, \beta_1, \beta_2)=(0.0002, 0.9, 0.999)$ with the cosine learning rate scheduling \cite{loshchilov2016sgdr}. We use 2K and 100K steps of a linear warm-up phase in learning rate for CIFAR and ImageNet models, respectively. Overall, we observe that a stable training of ViT (even for CIFAR-10) requires much stronger regularization compared to ResNets, otherwise they often significantly suffer from overfitting. In this respect, we apply the regularization practices those are now widely used for ViTs on ImageNet, namely mixup \cite{zhang2018mixup}, CutMix \cite{yun2019cutmix}, and RandAugment \cite{cubuk2020randaug}, following those established in \citet{beyer2022better}.

\subsection{Datasets}

\textbf{CIFAR-10/100} datasets \citep{dataset/cifar} consist of 60,000 images of size 32$\times$32 pixels, 50,000 for training and 10,000 for testing. Each of the images is labeled to one of 10 and 100 classes, and the number of data per class is set evenly, \ie 6,000 and 600 images per each class, respectively. By default, we use the random translation up to 4 pixels as a data pre-processing. 
We normalize the images in pixel-wise by the mean and the standard deviation calculated from the training set. The full dataset can be downloaded at \url{https://www.cs.toronto.edu/~kriz/cifar.html}.

{\textbf{CIFAR-10/100-C, and ImageNet-C} datasets \cite{hendrycks2018benchmarking} are collections of 75 replicas of the CIFAR-10/100 test datasets (of size 10,000) and ImageNet validation dataset (of size 50,000), respectively, which consists of 15 different types of common corruptions each of which contains 5 levels of corruption severities. Specifically, the datasets includes the following corruption types: (a) \emph{noise}: Gaussian, shot, and impulse noise; (b) \emph{blur}: defocus, glass, motion, zoom; (c) \emph{weather}: snow, frost, fog, bright; and (d) \emph{digital}: contrast, elastic, pixel, JPEG compression. In our experiments, we evaluate test errors on CIFAR-10/100-C for models trained on the ``clean'' CIFAR-10/100 datasets, where the error values are averaged across different corruption types per severity level. For ImageNet-C, on the other hand, we compute and compare the mean Corruption Error (mCE) proposed by \citet{hendrycks2018benchmarking}. Specifically, mCE is the average of Corruption Error (CE) over corruption types, where CE is defined by the error rates normalized by those from AlexNet \cite{kriz2012alexnet} to adjust varying difficulties across corruption types. Formally, for a classifier $f$, CE for a specific corruption type $c$ is defined by:
\begin{equation}
    \mathrm{CE}_c^f := \left(\sum_{s=1}^5 \mathtt{error}_{c, s}^f \right) \Bigg/ \left( \sum_{s=1}^5 \mathtt{error}_{c, s}^{\mathrm{AlexNet}} \right),
\end{equation}
where $s$ denotes the severity level ($1 \le s \le 5$). The full datasets, as well as the information on the pre-computed AlexNet error rates on ImageNet-C (to compute mCE), can be downloaded at \url{https://github.com/hendrycks/robustness}.}

{\textbf{CIFAR-10.1/10.2} datasets \cite{recht2018cifar10.1,lu2020harder} are reproductions of the CIFAR-10 test set that are separately collected from Tiny Images dataset \cite{torralba2008tiny}. Both datasets consist 2,000 samples for testing, and designed to minimize distribution shift relative to the original CIFAR-10 dataset in their data creation pipelines. The datasets can be downloaded at \url{https://github.com/modestyachts/CIFAR-10.1} (for CIFAR-10.1; we use the ``\texttt{v6}'' version) and \url{https://github.com/modestyachts/cifar-10.2} (for CIFAR-10.2).}

{\textbf{CINIC-10} dataset \cite{darlow2018cinic} is an extension of the CIFAR-10 dataset generated via addition of down-sampled ImageNet images. The dataset consists of 270,000 images in total of size 32$\times$32 pixels, those are equally distributed for train, validation and test splits, \ie the test dataset (that we use for our evaluation) consists of 90,000 samples. The full datasets can be downloaded at \url{https://github.com/BayesWatch/cinic-10}.}

{\textbf{ImageNet} dataset \cite{dataset/ilsvrc}, also known as ILSVRC 2012 classification dataset, consists of 1.2 million high-resolution training images and 50,000 validation images, which are labeled with 1,000 classes. As a data pre-processing step, we perform a 256$\times$256 resized random cropping and horizontal flipping for training images. For testing images, on the other hand, we apply a 256$\times$256 center cropping for testing images after re-scaling the images to have 256 in their shorter edges. Similar to CIFAR-10, all the images are normalized by the pre-computed mean and standard deviation. A link for downloading the full dataset can be found in \url{http://image-net.org/download}.}

{\textbf{ImageNet-R} dataset \cite{Hendrycks_2021_ICCV} consists of 30,000 images of various artistic renditions for 200 (out of 1,000) ImageNet classes: \eg art, cartoons, deviantart, graffiti, embroidery, graphics, origami, paintings, patterns, plastic objects, plush objects, sculptures, sketches, tattoos, toys, video game renditions, and so on. To perform an evaluation of ImageNet classifiers on this dataset, we apply masking on classifier logits for the 800 classes those are not in ImageNet-R. The full dataset can be downloaded at \url{https://github.com/hendrycks/imagenet-r}.}

{\textbf{ImageNet-Sketch} dataset \cite{wang2019learning} consists of 50,000 sketch-like images, 50 images for each of the 1,000 ImageNet classes. The dataset is constructed with Google image search, using queries of the form ``{\tt sketch of [CLS]}'' within the ``black and white'' color scheme, where {\tt [CLS]} is the placeholder for class names. The full dataset as well as the scripts to collect the dataset can be accessed at \url{https://github.com/HaohanWang/ImageNet-Sketch}.}

\textbf{CelebFaces Attributes (CelebA)} dataset \cite{liu2015faceattributes} consists of 202,599 face images, where each is labeled with 40 attribute annotations. We follow the standard train/validation/test splits of the dataset as provided by \citet{liu2015faceattributes}, and use the train split for training and computing FID scores following the protocol of other baselines \cite{parmar2021dual,aneja2021contrastive}. We also follow the pre-processing procedure of \cite{liu2015faceattributes} to fit in the images into the size of 64$\times$64: namely, we first perform a center crop into size 140$\times$140 to the images, followed by a resizing operation into 64$\times$64. The full dataset can be downloaded at \url{https://mmlab.ie.cuhk.edu.hk/projects/CelebA.html}.

\subsection{Computing infrastructure} 

Unless otherwise noted, we use a single NVIDIA Geforce RTX-2080Ti GPU to execute each of the experiments. For experiments based on StyleGAN2 architecture (reported in Table~\ref{table:csd_fid}), we use two NVIDIA Geforce RTX-2080Ti GPUs per run. For the ImageNet experiments (reported in Table~\ref{tab:imagenet} in the main text), we use 8 NVIDIA Geforce RTX-3090 GPUs per run. Overall, our experiments are based on well-established encoder architectures such as ResNet and ViT, and the model size of AENIB is generally followed by these backbone models.

\clearpage
\FloatBarrier
\section{Architectural details}
\label{appendix:architecture}

\begin{figure}[ht]
\begin{subfigure}{\linewidth}
    \centering
    \includegraphics[width=0.85\linewidth]{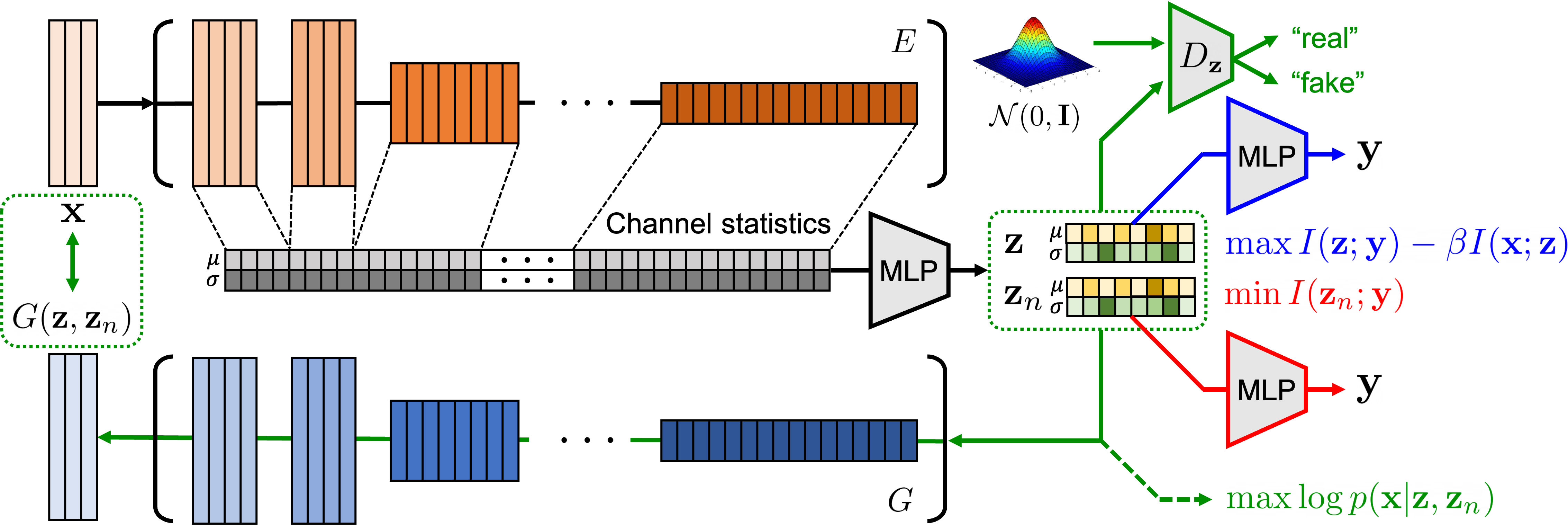}
    \caption{ConvNet-based AENIB}
\end{subfigure}
\begin{subfigure}{\linewidth}
    \centering
    \includegraphics[width=0.65\linewidth]{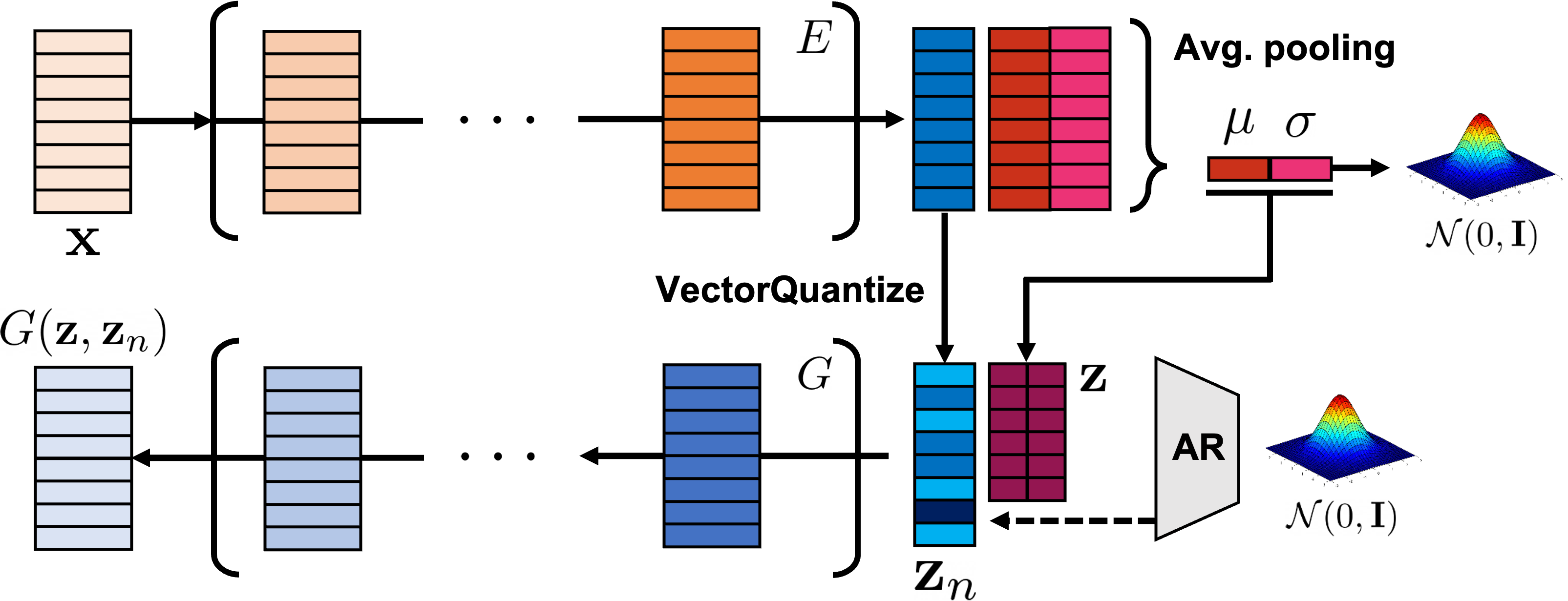}
    \caption{Transformer-based AENIB}
\end{subfigure}
\caption{An overview of the proposed framework, \emph{nuisance-extended information bottleneck} (NIB), instantiated by an autoencoder-based design with (a) ConvNet-based and 
 (b) Transformer-based architectures.}\label{figure:overall_vit}
\end{figure}

Recall that our proposed {{\method}} architecture consists of (a) an encoder $f$, (b) a decoder $g$, and (c) MLP-based discriminators $d_\rvy$, $d_\rvz$, and an MLP for feature statistic projection $\Pi_f$. We set 128 as the nuisance dimension $\rvz_n$, and use hidden layer of size 1,024 for MLP-based discriminators, \eg $d_\rvy$, $d_\rvz$, and MLPs for projection $\Pi_f$.

\subsection{ConvNet-based architecture}
\label{ap:arch_convnet}

We mainly consider ResNet-18 \citep{he2016deep} as a ConvNet-based encoder. For this encoder, we consider the generator architecture of FastGAN \citep{liu2021towards} as the decoder, but with a modification on normalization layers: specifically, we replace the standard batch normalization \cite{ioffe2015batch} layers in the architecture with adaptive instance normalization (AdaIN) \cite{karras2019style} so that the affine parameters can be modulated by $\rvz$ and $\rvz_n$ as well as the decoder input: we observe a consistent gain in FID from this modification.

\vspace{0.05in}
\noindent\textbf{Adversarial similarity based guidance. }
We present an additional \emph{adversarial objective} leveraging the efficiency of \emph{feature statistics} based encoder (see Section~\ref{ss:arch}) for ConvNet-based AENIB models to boost their generative modeling. Namely, we use the feature statistics based encoder to also provide the decoder $g$ an extra guidance in minimizing the (pixel-level) reconstruction loss \eqref{eq:recon}. We propose to additionally place a discriminator network, say $d_{\rvx}: \mathbb{R}^{|\Pi_E|}\rightarrow {\mathbb{R}}^{e}$, that computes similarity between $\Pi_f(\rvx)$ and $\Pi_f(g(\rvz, \rvz_n))$ and to perform an adversarial training: 
\begin{equation}\label{eq:adv_sim}
    L_{\tt sim} := \max_{d_{\rvx}}\ \log(1 - \sigma(\tfrac{1}{\tau} \cdot \mathrm{sim}(d_{\rvx}(\Pi_f(\rvx)), d_{\rvx}(\Pi_f(g(\rvz, \rvz_n)))))),
\end{equation}
where $\sigma(\cdot)$ is the sigmoid function and $\mathrm{sim}(\rvx, \rvy):=\tfrac{\rvx \cdot \rvy}{\|\rvx\|\|\rvy\|}$ denotes the cosine similarity. Here, $\tau$ is a temperature hyperparameter, and we use $\tau=0.2$ throughout our experiments. We apply this additional objective for ConvNet-based {\method} models in our experiments by default, which turns out to be helpful particularly for the generation quality of the learned autoencoders.

\subsection{Transformer-based architecture}

We consider ViT-S and ViT-B \citep{dosovitskiy2021an,touvron2021training} in our experiments. When Transformer-based encoder is used, we use the same Transformer architecture as the decoder model where it is preceded by linear layers that maps both $\rvz$ and $\rvz_n$ into the space of patch embedding. We assume the patch size of ViT to be 4 for CIFAR-10 and 16 for ImageNet, \ie we denote it as ViT-S/4 and ViT-S/16, respectively, so that the outputs from the models have similar numbers of patch embeddings ($8\times8$ and $16\times16$, respectively) to those of ResNet-18. To model $\rvz$ and $\rvz_n$ in the ViT architecture, we simply split the output patch embedding into two separate embeddings (of reduced embedding dimensions): one of these embeddings is average-pooled to define $\rvz$, and the remaining one is vector-quantized \cite{yu2022vectorquantized} to define a nuisance representation $\rvz_n$, as described in the next paragraph in more details. Figure~\ref{figure:overall_vit} illustrates an overview of our proposed {\method} for ViT-based architectures.
 
\vspace{0.05in}
\noindent\textbf{VQ-based nuisance modeling for ViT-{\method}. } Remark that our current ViT-based design allocates \emph{different} numbers of feature dimension for $\rvz$ and $\rvz_n$: specifically, we only apply global average pooling for $\rvz$ (not for $\rvz_n$), so that its dimensionality becomes independent to the input resolution, while the nuisance $\rvz_n$ would still get an increasing dimensionality for higher-resolution inputs. In practice, this difference in feature dimension may cause some training difficulties in {\method} training: (a) it makes harder to balance between the objectives given that {\method} is essentially a ``competition'' between two information channels, \ie $\rvz$ and $\rvz_n$; also, (b) it becomes increasingly difficult to force $\rvz_n$ to follow the independent Gaussian marginal for a tractable sampling as $\rvz_n$ gets higher dimensions, unlike our ConvNet-based design. To alleviate these issues, we propose to apply the \emph{vector quantization} (VQ) \cite{oord2017neural,yu2022vectorquantized} to the nuisance embedding: namely, we train the output of the nuisnace head, say $\hat{\rvz}_n$, to have one of (discrete) vectors in a learned dictionary $\mathbf{e}$. With this, now $\rvz_n$ becomes independent to the dimensionality of per-patch embeddings, which allows a more scalable balancing with $\rvz$, as well as offering a tractable marginal distribution to sample: given that $\rvz_n$ now gets a sequence of discrete distribution, one can apply a post-hoc generative modeling (\eg with an autoregressive prior \cite{oord2017neural}, or with a diffusion model \cite{gu2022vector}) to allow an efficient sampling. Specifically, we add the following objective upon our proposed {\method} objective \eqref{eq:final} to enable VQ-based nuisance modeling:
\begin{equation}
    L_{\tt VQ}(\rvx; \mathbf{e}) := \| \mathtt{sg}[\hat{\mathbf{z}}_n(\mathbf{x})] - e \|_2^2 + \beta\| \hat{\mathbf{z}}_n(\mathbf{x}) - \mathtt{sg}[e] \|_2^2,
\end{equation}
where $e := \min_i \|\hat{\mathbf{z}}_n(\mathbf{x}) - e_i \|_2^2$, and $\mathtt{sg}(\cdot)$ denotes the stop-gradient operator defined by $\mathtt{sg}(x) \equiv x$ and $\frac{d}{dx}\mathtt{sg}(x) \equiv 0$. Here, the commitment hyperparameter $\beta$ is set to 0.25 following \cite{oord2017neural,yu2022vectorquantized}. We allocate 32 per-patch dimensions for $\rvz_n$, with an embedding dictionary $\mathbf{e}$ of size 256. We adopt the embedding normalization \cite{yu2022vectorquantized} as we found it consistently improves the stability of VQ-based training.

\vspace{0.05in}
\noindent\textbf{SSIM-based $D_2^2$ reconstruction loss. } Recall that our proposed {\method} is based on minimizing reconstruction loss \eqref{eq:recon} to implement $I(\rvx; \rvz, \rvz_n)$ in NIB \eqref{eq:nib}. Although we introduce the \emph{normalized mean-squared error} (NMSE) as a default design choice, the choice may not be limited to that: here, we demonstrate a SSIM-based \cite{wang2004image} reconstruction loss as an alternative, and show its effectiveness on improving corruption robustness. Specifically, for a given pair of images\footnote{In practice, SSIM is often computed in per-patch basis for a sliding window of a certain kernel size, \eg 8. The values are then averaged to define the metric. In our experiments, we also follow this implementation.} $(x, y)$, the \emph{structural similarity index measure} (SSIM) defines a similarity metric between $x$ and $y$ considering differences in luminance (represented by $S_1$) and structures (represented by $S_2$): 
\begin{equation}\label{eq:ssim}
    \mathrm{SSIM}(x, y) = \frac{2\mu_x \mu_y + c_1}{\mu_x^2 + \mu_y^2 + c_1} \cdot \frac{2\sigma_{xy} + c_2}{\sigma_x^2 + \sigma_y^2 + c_2} =: S_1 \cdot S_2,
\end{equation}
where $c_1:=0.01^2$ and $c_2:=0.03^2$ are small constants for numerical stability, as well as to simulate the saturation effects of visual system under low luminance (and contrast) \cite{brunet2011mathematical}. Given that SSIM itself is not a distance metric (\eg it often allows the value to be negative), however, we instead consider the following modification of SSIM, the \emph{squared-$D_2$} ($D_2^2$), as our reconstruction loss, where $D_2$ is originally defined by \citet{brunet2011mathematical} that is shown to be a distance: 
\begin{equation}\label{eq:d2_loss}
    D_2^2(x, y):= \sqrt{(1 - S_1) + (1 - S_2)}^2 = {2 - S_1 - S_2}.
\end{equation}
In Table~\ref{tab:d2_loss}, we compare the effect of having different reconstruction losses in AENIB between the default choice of NMSE and $D_2^2$: the results on CIFAR-10/100-C with ViT-S/4 show that $D_2^2$-based reconstruction loss can reliable improve corruption robustness of the AENIB models over NMSE. This confirms that the choice of reconstruction loss impacts the final robustness of AENIB, and also suggests that a more perceptually-aligned similarity metric could possibly make the model less biased toward spurious features that are not necessary to build a robust representation.

In this respect, we adopt the $D_2^2$-based loss in AENIB for ViT-based models in our experiments: somewhat interestingly, we found the objective becomes much harder to be minimized for ConvNet-based models, where we keep the default choice of NMSE. This is possibly because that there can be a discrepancy between what ConvNets typically extract and those from a $D_2^2$-based reconstruction.

\begin{table*}[h]
\centering
\begin{adjustbox}{width=0.95\linewidth}
\begin{tabular}{c|c|ccccccccccccccc|c}
    \toprule
    AENIB (ViT-S/4) & Loss &
    \rotatebox{75}{Gaussian} & \rotatebox{75}{Shot} & \rotatebox{75}{Impulse} & \rotatebox{75}{Defocus} & \rotatebox{75}{Glass} & \rotatebox{75}{Motion} & \rotatebox{75}{Zoom} & \rotatebox{75}{Snow} & \rotatebox{75}{Frost} & \rotatebox{75}{Fog} & \rotatebox{75}{Brightness} & \rotatebox{75}{Contrast} & \rotatebox{75}{Elastic} & \rotatebox{75}{Pixelate} & \rotatebox{75}{JPEG} & \rotatebox{75}{Average} \\ 
    \midrule
    \multirow{ 2}{*}{CIFAR-10-C} & {NMSE} & 20.0 & 15.8 & 19.2 & \textbf{10.2} & \textbf{18.3} & \textbf{12.7} & \textbf{11.9} &9.76 & 10.0 & 11.0 & 6.33 & 11.3 & \textbf{10.5} & \textbf{13.4} & 14.7 & 13.0\\
    & \textbf{$D_2^2$} & \textbf{17.8} & \textbf{14.0} & \textbf{17.6} & \textbf{10.2} & 19.7 & 13.2 & 12.4 & \textbf{9.12} & \textbf{8.87} & \textbf{9.56} & \textbf{5.86} & \textbf{8.29} & 10.7 & 14.3 & \textbf{13.5} & \textbf{12.3}  \\
    \midrule
    \multirow{ 2}{*}{CIFAR-100-C} & {NMSE} & 48.2 & 42.8 & 43.2 & 32.4 & 48.6 & 36.1 & 35.6 & 31.4 & 32.3 & 35.2 & 25.3 & 33.7 & 33.2 & 36.2 & 40.9 & 36.9 \\
    & \textbf{$D_2^2$} & \textbf{45.7} & \textbf{40.1} & \textbf{42.1} & \textbf{32.2} & \textbf{47.9} & \textbf{35.8} & \textbf{35.1} & \textbf{31.1} & \textbf{31.7} & \textbf{34.1} & \textbf{24.4} & \textbf{31.5} & \textbf{32.4} & \textbf{34.5} & \textbf{38.4} & \textbf{35.8} \\
    \bottomrule
\end{tabular}
\end{adjustbox}
\caption{Comparison of average per-corruption error rates (\%; $\rda$) on CIFAR-10/100-C \cite{hendrycks2018benchmarking}. We use ViT-S/4 for this experiment. All the models reported here are trained via AENIB but with different reconstruction losses.}
\label{tab:d2_loss}
\end{table*}

\section{Ablation study}
\label{appendix:ablation}

\begin{figure*}[ht]
    \centering
    \hfill
    \begin{minipage}[b]{.45\textwidth}
      \centering
      \includegraphics[width=0.9\textwidth]{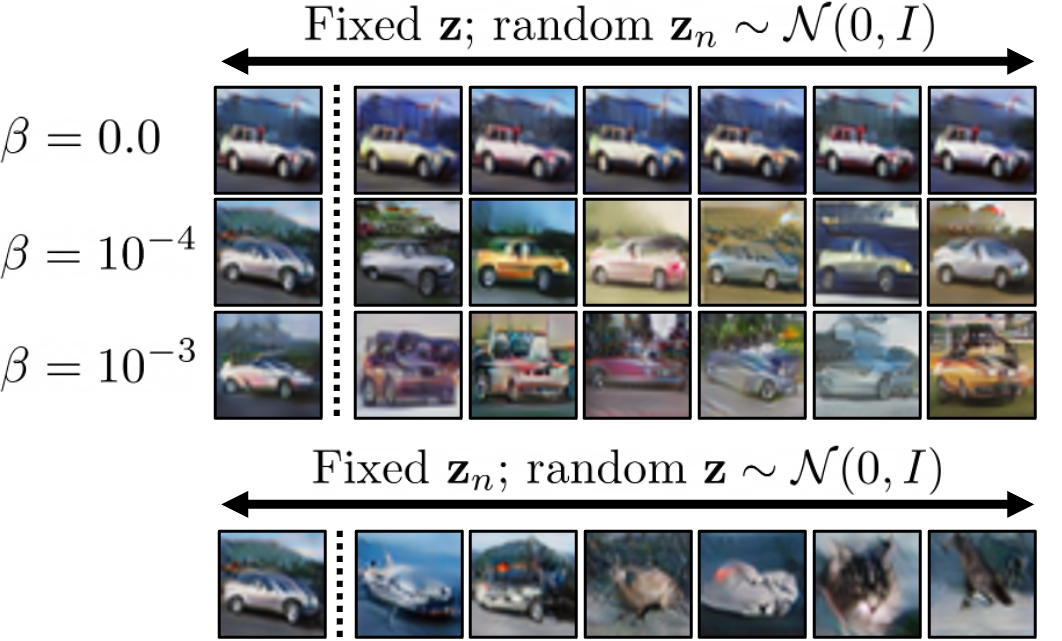}
      \captionof{figure}{Reconstructions under random nuisance $\rvz_n$ (upper rows) and random semantic $\rvz$ (lower row). The leftmost per row shows the original reconstruction.}
      \label{fig:resample}
    \end{minipage}
    \hfill
    \begin{minipage}[b]{0.48\linewidth}
    \begin{adjustbox}{width=\linewidth}
    \begin{tabular}{ccccc|ccc}
    \toprule
    $\beta$  & $L_{\tt recon}$  & {$L_{\tt nuis}$} & {$L_{\tt ind}$} & $L_{\tt sim}$  & Err. & {C-Err.} & FID  \\
    \midrule
    \tt{1e-4} & \ck & \ck & \ck & \ck & {7.07} & {{23.3}} & 33.3 \\ 
    \midrule
    \tt{1e-3} & \ck & \ck & \ck & \ck & 7.32 & {24.5} & 31.0 \\    
    \tt{1e-2} & \ck & \ck & \ck & \ck & 7.38 & {26.3} & 30.8 \\     
    \midrule
    \tt{1e-4} & \xk & \ck & \ck & \ck & 8.29 & {29.2} & 33.8 \\    
    \tt{1e-4} & \ck & \xk & \ck & \ck & {8.01} & {24.1} & {{29.2}}  \\  
    \tt{1e-4} & \ck & \ck & \xk & \ck & {7.31} & {22.4} & {{78.3}}  \\  
    \tt{1e-4} & \ck & \ck & \ck & \xk & 7.95 & {28.6} & 83.1 \\  %
    \bottomrule
\end{tabular}

    \end{adjustbox}
    \captionof{table}{Comparison of the test error rate (Err.; \%, $\rda$), corruption error (C-Err.; \%, $\rda$) and FID on CIFAR-10 across ablations. We use ResNet-18 architecture for this experiment.}
    \label{tab:ablation}
    \end{minipage}
    \hfill
\end{figure*}

We further perform an ablation study on CIFAR-10 for a detailed analysis of the proposed {{\method}}. We use ResNet-18 based AENIB throughout this section.

\vspace{0.05in}
\noindent\textbf{Effect of $\beta$. } As also introduced in the original IB objective, $\beta \ge 0$ plays the key role in {{\method}} training as it controls the information balance between the semantic $\rvz$ and the nuisance $\rvz_n$. Here, Figure~\ref{fig:resample} examine how using different value of $\beta$ affect the actual representations, by comparing the reconstructed samples for a fixed input while randomizing the nuisance $\rvz_n$. Indeed, we observe a clear trend from this comparison demonstrating the effect of $\beta$: having larger $\beta$ makes the model to push more ``semantic'' information into $\rvz_n$ regarding it as the nuisance. Without information bottleneck, \ie in case when $\beta=0.0$, we qualitatively observe that the network rather encodes most information in $\rvz$, due to the minimax loss applied to the nuisance $\rvz_n$. {Quantitatively, this behavior is further evidenced in Table~\ref{tab:ablation} as an increase in the corruption errors when using larger $\beta$.}

\vspace{0.05in}
\noindent\textbf{Reconstruction loss. } The reconstruction loss $L_{\tt recon}$ is one of essential part to make {{\method}} work as a ``nuisance modeling'': in Table~\ref{tab:ablation}, we provide an ablation when this loss is omitted, showing a significant degradation in the final accuracy, {and more crucially in the corruption error. This confirms the necessity of reconstruction loss to obtain a robust representation in {{\method}}.} Nevertheless, due to the adversarial similarity loss $L_{\tt sim}$ that can also work (while not perfectly) as a reconstruction loss, one can still observe that the FID of the model can be moderately preserved.  

\vspace{0.05in}
\noindent\textbf{Nuisance loss. } From the ablation of $L_{\tt nuis}$ given in Table~\ref{tab:ablation}, we observe not only a considerable degradation in clean accuracy but also in its corruption robustness. This shows that strictly forcing the nuisance-ness to $\rvz_n$ (against $\rvy$) indeed helps $\rvz$ to learn a more robust representation, possibly from encouraging $\rvz$ to extract more diverse class-related information in a faithful manner by keeping the remainder information in $\rvz_n$ sufficient to infer $\rvx$.

\vspace{0.05in}
\noindent\textbf{Independence loss. } 
The independence loss $L_{\tt ind}$ in our current design, which essentially performs a GAN training toward $p (\rvz, \rvz_n) \sim \mathcal{N}(0, \mathbf{I})$, not only forces $\rvz \perp \rvz_n$ but also leads $\rvz$ and $\rvz_n$ to have a tractable marginal distribution: so that one could efficiently perform a sampling from the learned decoder. In a practical aspect, therefore, omitting $L_{\tt ind}$ in {\method} can directly harm its generation quality as given in Table~\ref{tab:ablation}.
Nevertheless, it is still remarkable that the ablation could rather improve the corruption error: this suggests that our current design of forcing the full Gaussian may be restrictive. An alternative design for the future work could assume a weaker condition for $\rvz$ and $\rvz_n$, instead with a more sophisticated sampling to obtain a valid generative model from {\method}. 

\vspace{0.05in}
\noindent\textbf{Adversarial similarity. } As detailed in Appendix~\ref{ap:arch_convnet}, we introduce an additional adversarial reconstruction loss $L_{\tt sim}$ \eqref{eq:adv_sim} for cases when the backbone is convolutional. When the $L_{\tt sim}$ is omitted in the ConvNet-based AENIB training, we observe a significant degradation in both FID and accuracy, in a similar manner to $L_{\tt recon}$ but with a more impact on FID. This is reasonable given that $L_{\tt sim}$ and $L_{\tt recon}$ have the same goal of reconstructing input but in different metrics. Overall, it confirms the effectiveness of the proposed adversarial similarity based guidance to improve decoder performance.

\section{Proof of Lemma~1}
\label{appendix:proof}

\vspace{0.05in}
\noindent\textbf{Lemma~\ref{thm:noisy}.}
\textit{Let $\rvx \in \mathcal{X}$, and $\rvy \in \mathcal{Y}$ be random variables, $\hat{\rvx}$ be a noisy observation of $\rvx$ with $I(\rvx;\rvy) = I(\hat{\rvx};\rvy)$. Given that a representation $[\hat{\rvz}, \hat{\rvz}_n] := f(\hat{\rvx})$ of $\hat{\rvx}$ satisfies (a) $H(\hat{\rvx}|\hat{\rvz}, \hat{\rvz}_n)=0$, (b) $I(\hat{\rvz}_n; \rvy) = 0$, and (c) $\hat{\rvz} \perp \hat{\rvz}_n$, it holds $I(\hat{\rvz};\rvy) = I(\rvx;\rvy)$.}
\begin{proof}[Proof]
The condition $H(\hat{\rvx}|\hat{\rvz}, \hat{\rvz}_n)=0$ implies that there exists a deterministic $g$ such that $g(\hat{\rvz}, \hat{\rvz}_n) = \hat{\rvx}$ almost surely. Denoting $f(\cdot)$ by $f_0(\cdot, \rveps)$ with a certain deterministic mapping $f_0$ and an \textit{i.i.d.} random variable $\rveps \sim p(\rveps)$ via reparametrization trick, remark that a family of determinsitic functions $f_0^{\epsilon}(\cdot):= f_0(\cdot, \epsilon)$ becomes invertible almost surely for $\epsilon \sim p(\rveps)$ and $\hat{\rvx} \sim p(\hat{\rvx})$, \ie as $g(f_0^{\epsilon}(\hat{\rvx}))=\hat{\rvx}$. Then, the statement follows from the information preservation of invertible mapping and the chain rule of mutual information (and that of conditional mutual information), as well as by applying (b) and (c):
\begin{align}
    I(\rvx;\rvy) = I(\hat{\rvx};\rvy) &= I(\rvy; \hat{\rvz}, \hat{\rvz}_n) = I(\rvy; \hat{\rvz}_n) + I(\rvy; \hat{\rvz} | \hat{\rvz}_n) \\
    &= I(\rvy; \hat{\rvz}) + H(\hat{\rvz}_n| \rvy) + H(\hat{\rvz}_n| \hat{\rvz}) - H(\hat{\rvz}_n| \rvy, \hat{\rvz}) - H(\hat{\rvz}_n) \\
    &= I(\rvy; \hat{\rvz}) = I(\hat{\rvz}; \rvy).
\end{align}
\end{proof}

\section{Additional background}
\label{appendix:related}

\subsection{Detailed survey on related work}

\vspace{0.05in}
\noindent\textbf{Out-of-distribution robustness. } 
Since the seminal works \cite{szegedy2014intriguing,Nguyen_2015_CVPR,amodei2016concrete} revealing the fragility of neural networks for out-of-distribution inputs, there have been significant attempts on identifying and improving various notions of robustness: \eg detecting novel inputs \cite{hendrycks2017a,lee2018maha,hendrycks2018deep,hendrycks2019using,lee2018training,tack2020csi,xiao2020lregret}, robustness against corruptions \cite{hendrycks2018benchmarking,geirhos2018imagenettrained,hendrycks2020augmix,wang2021tent,diffenderfer2021winning}, and adversarial noise \cite{goodfellow2014explaining,madry2018towards,pmlr-v80-athalye18a,zhang2019theoretically,cohen2019certified,carlini2019evaluating}, to name a few. Due to its fundamental challenges in making neural network to extrapolate, however, most of the advances in the robustness literature has been made under assuming priors closely related to the individual problems: \eg \emph{Outlier Exposure} \cite{hendrycks2018deep} and \emph{AugMix} \cite{hendrycks2020augmix} assume an external dataset or a pipeline of data augmentations to improve the performances in novelty detection and corruption robustness, respectively; \emph{Tent} \cite{wang2021tent} leverages extra information available from a batch of samples in test-time to adapt a given neural network; \citet{tramer2019adversarial,kang2019transfer} observe that neural networks robust to a certain type of adversarial attack (\eg an $\ell_\infty$-constrained adversary) do not necessarily robust to other types of adversary (\eg an $\ell_1$ adversary), \ie adversarial robustness hardly generalizes from the adversary assumed \emph{a priori} for training. In this work, we aim to improve multiple notions of robustness without assuming such priors, through a new training scheme that extends the standard information bottleneck principle under noisy observations. 

\vspace{0.05in}
\noindent\textbf{Hybrid generative-discriminative modeling. }
Our proposed method can be also viewed as a new approach of improving the robustness of discriminative models by incorporating a generative model, in the context that has been explored in recent works \cite{lee2018maha,schott2018towards,Grathwohl2020Your,Yang_2021_ICCV}: for example, \citet{lee2018maha, pmlr-v97-lee19f} have shown that assuming a simple Gaussian mixture model on the deep discriminative representations can improve novelty detection and robustness to noisy labels, respectively; \citet{schott2018towards} develop an empirical defense against adversarial examples via generative classifiers; A line of research on \emph{Joint Energy-based Models} (JEM) \cite{Grathwohl2020Your,Yang_2021_ICCV} assumes the entire discriminative model as a joint generative model by interpreting the logits of $p(\rvy|\rvx)$ as unnormalized log-densities of $p(\rvx|\rvy)$, and shows that modeling $p(\rvx|\rvy)$ as well as $p(\rvy|\rvx)$ can improve out-of-distribution generalization of the classifier. 
Nevertheless, it is still an unexplored and open question that how to ``better'' incorporate generative representation into  discriminative models: in case of novelty detection, for example, several recent works \cite{nalisnick2018do, ren2019lratio, Serra2020Input, xiao2020lregret} observe that existing likelihood-based generative models are not accurate enough to detect out-of-distribution datasets, suggesting that relying solely on (likelihood-based) deep generative representation may not enough for robust classification \cite{Fetaya2020Understanding}. In case of JEM, on the other hand, it has been shown that directly assuming a joint generative-discriminative representation often makes a significant training instability. In this work, we propose to introduce an autoencoder-based model to avoid the training instability, and consider a design that the \emph{nuisance} can succinctly supplement the given discriminative representation to be generative.

\vspace{0.05in}
\noindent\textbf{Invertible representations {and nuisance modeling}. } The idea of incorporating nuisances can be also considered in the context of \emph{invertible} modeling, or as known as \emph{flow-based models} \cite{dinh2016density,kingma2018glow,jacobsen2018irevnet,behrmann2019invertible,chen2019residual,grathwohl2019ffjord},\footnote{A more complete survey on flow-based models can be found in \cite{kobyzev2020normalizing}.} which maps a given input $\rvx$ into a representation $\rvz$ of the same dimension so that one can construct an inverse of $\rvz$ to $\rvx$: here, the nuisance can be naturally defined as the remainder information of $\rvz$ for a given subspace of interest, \eg to model $\rvy$. For example, \citet{jacobsen2018excessive} adopt a fully-invertible variant of i-RevNet \cite{jacobsen2018irevnet} to analyze \emph{excessive invariance} in neural networks, \ie the existence of pairs of completely different samples with the same representation in a neural network, and proposes to maximize the cross-entropy for the nuisances in a similar manner to our proposed minimax-based nuisance loss (\eqref{eq:nuis} in the main text); \citet{lynton2020training}, on the other hand, leverages invertible neural network to model a Gaussian mixture based generative classifier in the representation space, so that nuisance information can be preserved until its representation. 
Compared to such approaches relying on invertible neural networks, our autoencoder-based nuisance modeling does not guarantee the ``full'' invertibility for arbitrary inputs: instead, it only focuses on inverting the data manifold given, and this enables (a) a much flexible encoder design in practice, \ie other than flow-based designs, and (b) a more scalable generative modeling of nuisance representation $\rvz_n$ while forcing its \emph{independence} to the semantic space $\rvz$. {This is due to that it works on a compact space rather than those proportional to the input dimension, which is an important benefit of our modeling in terms of the scalability of nuisance-aware training, \eg beyond an MNIST-scale as done by \citet{jacobsen2018excessive}.}
{More closer related works \cite{jaiswal2018unsupervised, jaiswal2019discovery, pan2021disentangled} in this respect instead introduce a separate encoder for nuisance factors, where the nuisanceness is induced by the independence to $\rvz$: \eg DisenIB \cite{pan2021disentangled} applies FactorVAE \cite{kim2018disentangling} between semantic and nuisance embeddings to force their independence.\footnote{{We provide a more direct empirical comparison with DisenIB \cite{pan2021disentangled} to {\method} in Section~\ref{exp:disenib}.}} Yet, similarly to the invertible approach, the literature has been questioned on that the idea can be scaled-up beyond, \eg MNIST, and our work does explore and establish a practical design that is applicable for recent architectures and datasets addressing modern security metrics, \eg corruption robustness. On the technical side, we find that, \eg the ``nuisanceness to $\rvy$'' is more important for $\rvz_n$ than the ``independence with $\rvz$'' (as usually done in the previous works \cite{jaiswal2018unsupervised, jaiswal2019discovery, pan2021disentangled}) to induce a robust representation, as verified in our ablation study in Appendix~\ref{appendix:ablation}, which can be a useful practice for the future research concerning robust representation learning.} 

\vspace{0.05in}
\noindent\textbf{Autoencoder-based generative models. } 
There have been steady advances in generative modeling based on autoencoder architectures, especially since the development in \emph{variational autoencoders} (VAEs) \cite{kingma2014autoencoding}: due to its ability of estimating data likelihoods, and its flexibility to implement various statistical assumptions \cite{louizos2015variational,kingma2016improved,kim2018disentangling}. With the advances in its training objectives \cite{pascal2008denoising,makhzani2015adversarial,higgins2016beta} as well as the architectural improvements \cite{vahdat2020nvae,child2021very}, VAE-based models are currently considered as one of state-of-the-art approaches in likelihood based generative modeling: \eg a state-of-the-art \emph{diffusion models}~\citep{ho2020ddpm, song2021denoising} is built upon the denoising autoencoders under Gaussian perturbations, and recently-proposed \emph{hierarchical VAEs} \cite{vahdat2020nvae,child2021very} have shown that VAEs can benefit from scaling up its architectures into deeper encoder networks. In perspectives of viewing our method as a \emph{generative modeling}, {{\method}} is based on \emph{adversarial autoencoders} \cite{makhzani2015adversarial} that replaces the KL-divergence based regularization in standard VAEs with a GAN-based adversarial loss, with a novel encoder architecture that is based on the \emph{internal feature statistics} of discriminative models: so that the model can better encode lower-level features without changing the backbone architecture. We observe that this design enables autoencoder-based modeling even from a large, pre-trained discriminative models, and this ``projection'' of internal features can significantly benefit the generation quality, as well as for generative adversarial networks (GANs) as observed in Table~\ref{table:csd_fid}.

\subsection{Technical background}
\label{appendix:tech}

\vspace{0.05in}
\noindent\textbf{Variational information bottleneck. }
Although the information bottleneck (IB) principle given in \eqref{eq:ib} \cite{tishby99information} suggests a useful definition on what we mean by a ``good'' representation, computing mutual information of two random variables is generally hard and this makes the IB objective infeasible in practice. To overcome this, \emph{variational information bottleneck} (VIB) \cite{alemi2016deep,chalk2016relevant} applies variational inference to obtain a lower bound on the IB objective \eqref{eq:ib}. Specifically, it approximates: (a) $p(\rvy|\rvz)$ by a (parametrized) ``decoder'' neural network $q(\rvy|\rvz)$, and (b) $p(\rvz)$ by an ``easier'' distribution $r(\rvz)$, \eg isotropic Gaussian $\mathcal{N}(\rvz|0, \mathbf{I})$. Having such (variational) approximations in computing \eqref{eq:ib} as well as the Markov chain property $\rvy - \rvx - \rvz$ of neural networks, one yields the following lower bound on the IB objective \eqref{eq:ib}:
\begin{align}\label{eq:vib}
    I(\rvz;\rvy) - \beta I(\rvz, \rvx) &\ge \mathbb{E}_{\rvx, \rvy}\left[\int dz \left(p(z|\rvx) \log q(\rvy|z) - \beta p(z|\rvx) \log \frac{p(z|\rvx)}{r(z)} \right) \right].
\end{align}
This bound can now be approximated with the empirical distribution {$p(\rvx, \rvy)\approx \tfrac{1}{n}\sum_i \delta_{x_i}(\rvx)\delta_{y_i}(\rvy)$} from data.
By further assuming a Gaussian encoder $p(\rvz|\rvx) := \mathcal{N}(\rvz|f^{\mu}(\rvx), f^{\sigma}(\rvx))$ as defined in \eqref{eq:gaussian_decoder} and applying the reprarametrization trick \cite{kingma2014autoencoding}, we get the following VIB objective:
\begin{equation}
    L_{\tt VIB}^{\beta} := \frac{1}{n} \sum_{i=1}^{n}\ \mathbb{E}_{\rveps}[-\log q(y_i | f(x_i, \rveps))] + \beta\  \mathrm{KL}~(p(\rvz|x_i) \| r(\rvz)).
\end{equation}

\vspace{0.05in}
\noindent\textbf{Generative adversarial networks. }
\emph{Generative adversarial network} (GAN) \cite{goodfellow2014gan} considers the problem of learning a generative model $p_g$ from given data $\{x_i\}_{i=1}^n$, where $x_i\sim p_d(\rvx)$ and $\rvx \in \mathcal{X}$. Specifically, GAN consists of two neural networks: (a) a \emph{generator} network $G: \mathcal{Z} \rightarrow \mathcal{X}$ that maps a latent variable $z \sim p(\rvz)$ into $\mathcal{X}$, where $p(\rvz)$ is a specific prior distribution, and (b) a \emph{discriminator} network $D: \mathcal{X} \rightarrow [0, 1]$ that discriminates samples from $p_d$ and those from the implicit distribution $p_g$ derived from $G(\rvz)$. The primitive form of training $G$ and $D$ is the following:
\begin{equation}\label{eq:minimax}
    \min_{G}\max_{D} V(G, D) \coloneqq \mathbb{E}_{\rvx}[\log (D(\rvx))] 
        + \mathbb{E}_{\rvz}[\log (1 - D(G(\rvz)))].
\end{equation}
For a fixed $G$, the inner maximization objective \eqref{eq:minimax} with respect to $D$ leads to the following \emph{optimal discriminator} $D^*_G$, and consequently the outer minimization objective with respect to $G$ becomes to minimize the \emph{Jensen-Shannon divergence} between $p_d$ and $p_g$, namely $D^*_{G} := \frac{p_d}{p_d+p_g}$.

\FloatBarrier
\section{Results on MNIST-C}
\label{appendix:mnistc}

\begin{table*}[ht]
\centering
\small
\begin{adjustbox}{width=\linewidth}
\setlength{\tabcolsep}{3pt}
\begin{tabular}{l|c|c|ccccccccccccccc|c}
    \toprule
    Method & \rotatebox{75}{Clean} & \rotatebox{75}{AUROC ($\gua$)} & \rotatebox{75}{Shot} & \rotatebox{75}{Impulse} & \rotatebox{75}{Glass} & \rotatebox{75}{Motion} & \rotatebox{75}{Shear} & \rotatebox{75}{Scale} & \rotatebox{75}{Rotate} & \rotatebox{75}{Brightness} & \rotatebox{75}{Translate} & \rotatebox{75}{Stripe} & \rotatebox{75}{Fog} & \rotatebox{75}{Spatter} & \rotatebox{75}{Dotted line} & \rotatebox{75}{Zigzag} & \rotatebox{75}{Canny edges} & \rotatebox{75}{Average} \\ 
    \midrule
    {Cross-entropy} & 0.45 & 0.987 & 4.69 & 69.6 & 60.3 & 46.5 & 1.41 & 2.97 & \textbf{4.80} & 88.7 & 2.45 & 76.6 & 88.7 & 27.3 & 5.64 & \textbf{27.3} & 44.1 & 34.5 \\
    {VIB \cite{alemi2016deep}
    } & \textbf{0.44} & 0.988 & 4.52 & 73.5 & 73.8 & 71.8 & 1.73 & 2.84 & 5.85 & 90.1 & 2.15 & 78.1 & 89.8 & 28.4 & 5.85 & 28.5 & 44.0 & 37.6 \\
    {sq-VIB \cite{thobaben2020convex}
    } & 0.48 & 0.955 & 4.32 & 71.5 & 63.5 & 62.3 & 1.62 & 2.70 & 5.74 & 90.5 & 2.43 & 80.3 & 90.3 & 24.8 & 5.91 & 32.0 & 43.4 & 36.4 \\
    {NLIB \cite{kolchinsky2019nonlinear}
    } & 1.15 & 0.974 & 7.13 & 67.9 & 62.5 & 57.9 & 2.15 & 4.00 & 7.06 & \textbf{86.9} & 3.28 & 81.8 & 88.7 & 30.1 & 8.97 & 31.0 & 41.8 & 36.4 \\
    {sq-NLIB \cite{thobaben2020convex}
    } & 3.19 & 0.908 & 9.90 & 73.3 & 66.7 & 64.7 & 4.25 & 6.19 & 9.21 & 88.7 & 6.43 & \textbf{72.4} & 89.8 & 32.4 & 9.69 & 36.2 & 72.5 & 40.3 \\
    {DisenIB \cite{pan2021disentangled}
    } & 0.54 & 0.997 & 4.60 & 68.8 & 56.4 & 50.4 & 1.11 & \textbf{2.04} & 4.84 & 88.7 & 2.01 & 74.3 & \textbf{88.5} & 20.1 & 4.75 & 27.4 & 69.0 & 35.2 \\
    \midrule
    {{\textbf{{\method} (ours)}}} & {0.72} & \textbf{1.000} & \textbf{3.71} & \textcolor{blue}{\textbf{48.8}} & \textcolor{blue}{\textbf{44.0}} & \textcolor{blue}{\textbf{27.1}} & \textbf{0.99} & {3.15} & {4.82} & {89.7} & \textbf{0.88} & {82.0} & {89.7} & \textcolor{blue}{\textbf{16.4}} & \textbf{4.14} & {33.9} & \textcolor{blue}{\textbf{25.9}} & \textcolor{blue}{\textbf{29.8}} \\
    \bottomrule
\end{tabular}

\end{adjustbox}
\caption{{Comparison of (a) clean error (\%; $\rda$), (b) AUROC ($\gua$) on detecting Gaussian noise (higher is better), and (c) corruption errors (\%; $\rda$) per corruption type on MNIST-C \cite{mu2019mnist}. Each classifier is trained on MNIST with random translation as augmentation. We highlight our results as blue whenever the value improves the baselines more than 3\% in absolute values.}}
\label{tab:mnistc}
\end{table*}

We also evaluate our proposed {{\method}} training on MNIST-C \cite{mu2019mnist}, a collection of corrupted versions of the MNIST \cite{dataset/mnist} test dataset of 15 corruption types constructed in a similar manner to CIFAR-10/100-C \cite{hendrycks2018benchmarking}, to get a clearer view on the effectiveness of our method on a simpler setup. For this experiments, we use a simple 4-layer convolutional network (with batch normalization \cite{ioffe2015batch}) as the encoder architecture, and trained every model on the (clean) MNIST training dataset for 100K updates following other training details of the CIFAR experiments (see Appendix~\ref{appendix:setup:detail}): again, we notice that the training does not assume specific prior on the corruptions. 
We compare {{\method}} with the direct ablations of cross-entropy and VIB based models, as well as some variants of VIB, namely Nonlinear-VIB \cite{kolchinsky2019nonlinear}, Squared-VIB/NLIB \cite{thobaben2020convex}, and DisenIB \cite{pan2021disentangled}. 

Table~\ref{tab:mnistc} summarizes the results: overall, we observe that the effectiveness of {{\method}} training still applies to MNIST-C, \eg our {{\method}} training improves the average corruption error from the baseline cross-entropy based training from $34.5\% \rightarrow 29.8\%$, which could not be obtained by simply sweeping on the baseline VIB training. 
Given that MNIST-C allows a visually clearer distinction between contents and corruptions compared to CIFAR-10/100-C, one can better interpret the behavior of given models on each corruption types: here, we observe that our training can dramatically improve robustness for certain types of corruptions where the baselines shows poor performances, \eg Impulse, Glass, and Motion, while still some types of corruptions are still remaining challenging even with {{\method}}, \eg especially for low-frequency biased corruptions such as Brightness and Stripe.
{Compared to DisenIB, on the other hand, we observe that the effectiveness from DisenIB, e.g., its gain in AUROC (as conducted by \citet{pan2021disentangled}), could not be further generalized on MNIST-C, where {{\method}} still improves upon it as well as achieving the perfect score at the same OOD task.}

\FloatBarrier
\section{Additional results on corruption robustness}
\label{appendix:additional_exp}

\begin{table*}[h]
\centering
\small
\begin{adjustbox}{width=\linewidth}
\begin{tabular}{l cccccc|c|cccccc|c}
    \toprule
     & \multicolumn{7}{c}{CIFAR-10-C} & \multicolumn{7}{c}{CIFAR-100-C} \\
    \cmidrule(r){2-8} \cmidrule(r){9-15}
         \multicolumn{1}{r}{Severity} & Clean & 1 & 2 & 3 & 4 & 5 & Avg. & Clean & 1 & 2 & 3 & 4 & 5 & Avg. \\
    \midrule
    Cross-entropy & \underline{5.71} & \underline{12.9} & 18.1 & 24.3	& 31.7	& 43.5 & {26.1} & \underline{26.9} & \underline{39.2} & \underline{46.9} & \underline{53.2} & \underline{59.8} & \underline{69.3} & \underline{53.7} \\
    VIB \cite{alemi2016deep} 
    & \textbf{5.47} & \textbf{12.5} & \underline{17.5} & \underline{23.6} & \underline{30.7} & \underline{42.5} & \underline{25.4} & \textbf{26.5} & 39.7 & 47.5 & 53.8 & 60.5 & 70.1 & 54.3 \\
    \cmidrule(r){1-1} \cmidrule{2-2} \cmidrule(lr){3-8} \cmidrule(lr){9-9} \cmidrule(l){10-15}
    {\textbf{{\method} (ours)}} & 7.07 & {13.2} & \textbf{17.2} & \textbf{21.7} & \textbf{27.5} & \textbf{37.0}
    & \textbf{23.3} & 28.0 & \textbf{39.0} & \textbf{45.5} & \textbf{51.4} & \textbf{57.6} & \textbf{67.0} & \textbf{52.1} \\
    \bottomrule
\end{tabular}
\end{adjustbox}
\caption{Comparison of average corruption error rates (\%; $\rda$) per severity level on \mbox{CIFAR-10/100-C} \cite{hendrycks2018benchmarking} with ResNet-18 architecture. Bold and underline denote the best and runner-up, respectively.  All the models are trained only using random translation as data augmentation.}
\label{tab:corr_resnet}
\end{table*}

Table~\ref{tab:corr_resnet} summarizes our experiments on corruption robustness with ResNet-18 architectures. Again, our AENIB consistently improves the robustness at common corruptions benefiting from nuisance modeling compared to VIB objectives. For example, we obtain an absolute $6.5\%$ of reduction in error rates on CIFAR-10-C of the highest severity, \ie by $43.5\% \rightarrow 37.0\%$. Here, we highlight that we do not utilize any data augmentations but random translation for training ResNet-18 models; the gain from our method indeed comes from the better robustness itself, unlike common practices in improving the corruption robustness by carefully designing augmentations. Interestingly, we observe that the impact of {{\method}} in the clean error can be different depending on the encoder architecture: compared to ViT results as given Table~\ref{tab:corruption_vit}, {{\method}} with ResNet-18 makes a slight decrease in clean accuracy, while the robust accuracy is still improved. In this case, it is more clear to see that AENIB could improve the \emph{effective robustness} of the learned representation. This is possibly due to that the representation induced via {{\method}} can be extracted better with non-local (attention-based) operations such as ViT.

\FloatBarrier
\section{{Experiments on image generation}}
\label{appendix:generation}

\begin{table*}[h]
    \centering
    \hfill
    \begin{minipage}[b]{0.5\linewidth}
        \centering
        \begin{adjustbox}{width=\linewidth}
        \begin{tabular}{lccc}
    \toprule
    CIFAR-10, Unconditional    & Augment.         &  FID ($\downarrow$)  &  {IS ($\uparrow$)}  \\ \midrule
    StyleGAN2 \cite{karras2020analyzing}    
    & HFlip &   {11.1}\phantom{*}  & 9.18 \\ 
    + DiffAug \cite{zhao2020diffaugment}    
    & Trans, CutOut &   9.89\phantom{*} & 9.40  \\
    + ContraD \cite{jeong2021training}      
    & SimCLR & 9.80\phantom{*} & 9.47  \\
    + ADA$^*$ \cite{karras2020training}   
    & Dynamic &  \textbf{7.01}$^*$ & -  \\
    \textbf{+ FSD (R-18; ours)} & {HFlip, Trans} &  {8.43}\phantom{*} & {9.68} \\ 
    \textbf{+ FSD (R-50; ours)} & {HFlip, Trans} &  \underline{7.39}\phantom{*} & \textbf{10.0}  \\ \midrule
    FastGAN \cite{liu2021towards}    
    & HFlip, Trans &   {34.5}\phantom{*}  & 6.52 \\ 
    + Proj-GAN (R-18) 
    & HFlip, Trans & 8.48\phantom{*} & 9.40  \\
    \textbf{+ FSD (R-18; ours)}   & {HFlip, Trans} & \textbf{7.80}\phantom{*} & \textbf{9.65}  \\
    \bottomrule
\end{tabular}
        \end{adjustbox}
        \captionof{table}{{Test FID and IS of GANs on CIFAR-10. Bold and underline indicate the best and runner-up, respectively. We note that the value of ADA$^*$ \cite{karras2020training} is taken after $2\times$ longer training steps.}}
        \label{table:csd_fid}
    \end{minipage}
    \hfill
    \begin{minipage}[b]{0.38\linewidth}
    \begin{adjustbox}{width=\linewidth}
    \begin{tabular}{lccc}
    \toprule
      & \multicolumn{2}{c}{CIFAR-10} & \multicolumn{1}{c}{CelebA} \\
    \cmidrule{2-3} \cmidrule(l){4-4} 
    \multicolumn{1}{c}{Method} & FID $\downarrow$ & IS $\uparrow$ & FID $\downarrow$ \\
    \midrule
     VAE \cite{parmar2021dual}  
     & 115.8 & 3.8 & -  \\
     VAE/GAN \cite{parmar2021dual} 
     & 39.8 & 7.4 & -  \\
     2s-VAE \cite{dai2018diagnosing} 
     & 72.9 & - & 44.4  \\
     Perceptual AE \cite{zhang2020perceptual} 
     & 51.5 & - & 13.8  \\
     NCP-VAE \cite{aneja2021contrastive} 
     & 24.1 & - & \textbf{5.25}  \\
     NVAE \cite{vahdat2020nvae} 
     & 56.0 & 5.19 & 13.5   \\
     DC-VAE \cite{parmar2021dual} 
     & \underline{17.9} & \underline{8.2} & 19.9   \\
     \midrule
     $L_{\tt recon}$ \eqref{eq:recon} only & {65.0} & {5.73}  & {50.1} \\
     {+ Adv. similarity} \eqref{eq:adv_sim} & {46.8} & {6.29}  & {25.1} \\
     {+ Projection (R-18)} & \textbf{{12.6}} & \textbf{{8.86}}  & \underline{6.91} \\
    \bottomrule
\end{tabular}

    \end{adjustbox}
    \captionof{table}{Test FID and IS of VAE models on unconditional generation of CIFAR-10 and CelebA. {Bold and underline denote the best and runner-up, respectively}.} 
    \label{tab:fid}
    \end{minipage}
    \hfill
    \vspace{-0.1in}
\end{table*}

\subsection{{Feature statistics discriminator for GANs}}
\label{appendix:exp_fsd}

Designing a stable discriminator has been crucial for GANs: a usual practice in the literature is to have a separate, carefully-designed network with a comparable generator, but with a significant overhead. We observe that the \emph{internal feature statistics} of a convolutional encoder $f$ can be a surprisingly effective representation to define a simple yet efficient discriminator. Concretely, for a given encoder $f$ and an input $\rvx$, we consider $L$ intermediate feature maps of $\rvx$, namely $\rvx^{(1)}, \cdots, \rvx^{(L)}$ from $f(\rvx)$, 
{the \emph{features statistics discriminator} (FSD) we consider here is then simply a 3-layer MLP applied on $\Pi_{f}(\rvx)$ \eqref{eq:projection}. In the following, we empirically confirm that this simplest design of discriminator can dramatically accelerate GAN training, particularly when applied upon pre-trained discriminative encoders: similarly to \citet{sauer2021projected} but with a simpler architecture.}

We evaluate the effect of the proposed feature statistics discriminator to the generation quality of GANs: here, we consider ImageNet-pretrained ResNet-18 (R-18) and ResNet-50 (R-50) \cite{he2016deep}, and define GAN discriminators via FSD upon the pre-trained models. We adopt StyleGAN2 \cite{karras2020analyzing} and FastGAN \cite{liu2021towards} for the generator architectures. For the StyleGAN2-based models, we follow the training details of DiffAug \cite{zhao2020diffaugment} and ADA \cite{karras2020training} in their CIFAR experiments: specifically, we use Adam with $(\alpha, \beta_1, \beta_2)=(0.002, 0.0, 0.99)$ for optimization with batch size of 64. We use non-saturating loss for training, and use $R_1$ regularization \cite{mescheder2018training} with $\gamma=0.01$. We do not use, however, the path length regularization and the lazy regularization \cite{karras2020analyzing} in training. We take exponential moving average on the generator weights with half-life of 500K samples. We stop training after 800K generator updates, which is about the half of those conducted for the ADA baseline \cite{karras2020training}. For the FastGAN baseline, on the other hand, we run the official implementation of FastGAN\footnote{\url{https://github.com/odegeasslbc/FastGAN-pytorch}} \cite{liu2021towards} on CIFAR-10 for the length of 6.4M samples with batch size 16. For the ``Projected GAN'' baseline, we adapt the official implementation\footnote{\url{https://github.com/autonomousvision/projected_gan}} \cite{sauer2021projected} onto the ImageNet pre-trained ResNet-18, and trained for 6.4M samples with batch size 64. Our results (``FSD'') follows the same training details, but with a difference in its discriminator. 

Table~\ref{table:csd_fid} summarizes the results. Overall, we observe that FSD can aid GAN training of given generator network surprisingly effectively: by leveraging pre-trained representations, FSD could achieve FID competitive with a state-of-the-art level approach of ADA \cite{karras2020training} even with using much weaker data augmentation. Compared to Projected GAN \cite{sauer2021projected} that also leverages pre-trained models to stabilize GANs, our approach offers a more simpler approach to leverage the given representations, \ie by just aggregating the features statistics, yet achieving a better FID. 

\subsection{Feature statistics encoder for autoencoders}
\label{exp:generation}

We also evaluate our proposed architecture and method as a \emph{generative modeling}, especially focusing on the effectiveness of the \emph{feature statistics encoder} (Section~\ref{ss:arch}) and the \emph{adversarial similarity} based training for ConvNet-based AENIB autoencoders on CIFAR-10 \cite{dataset/cifar} and CelebA \cite{liu2015faceattributes} datasets. To this end, we consider an ``unsupervised'' version of {{\method}} which omits the VIB loss ($L_{\tt VIB}^\beta$; \eqref{eq:vib}) and the nuisance loss ($L_{\tt nuis}$; \eqref{eq:nuis}) in training, so that the model can assume the setup of unconditional generation.

{Table~\ref{tab:fid} summarizes the quantitative generation results of our AENIB models optimized with different objectives.\footnote{Following other baselines, we compute FIDs from 50,000 generated samples against the training dataset.}} Firstly, it confirms the effectiveness of adversarial similarity based training: when it is solely applied upon $L_{\tt recon}$ (``$L_{\tt recon}$ only''; equivalent to \cite{makhzani2015adversarial}) it makes a significant improvements in both FID and IS. 
To further investigate the effectiveness of our proposed feature statistics encoder, we also test a scenario that the encoder is \emph{fixed} by \mbox{ResNet-18} pre-trained on ImageNet, akin to the setup of Table~\ref{table:csd_fid}: we observe that our encoder design can surprisingly benefit from using better representation, \eg ``+ Projection (R-18)'' in Table~\ref{tab:fid} further improves FID on CIFAR-10 from $46.8 \rightarrow 12.6$, better than the best results among considered VAE-based models, by only training an MLP upon the feature statistics of the (fixed) model. It is notable that the gain only appears when we apply the adversarial similarity based training: \ie even with the pre-trained model, it only achieves 67.5 in FID on CIFAR-10 without the training. This observation suggests an interesting direction to scale-up autoencoder-based models with large pre-trained representations, in a similar vein as \cite{sauer2021projected} as presented in the context of GANs. 

\clearpage
\FloatBarrier
\subsection{{Qualitative results}}

\begin{figure*}[ht]
  \centering
  \includegraphics[width=0.85\linewidth]{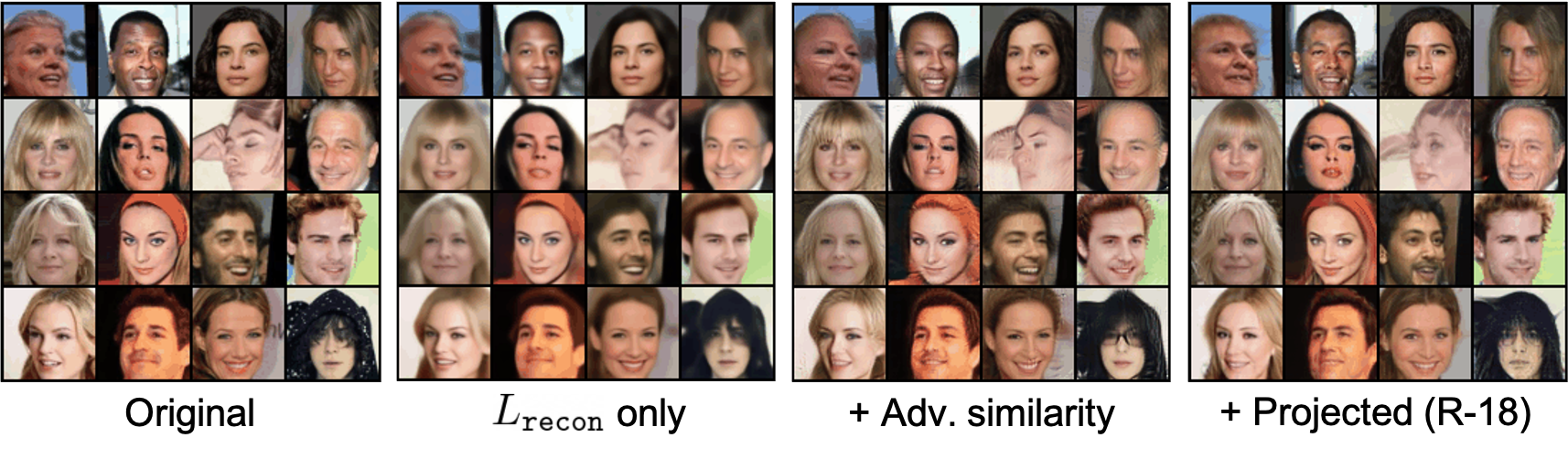}
 \vspace{-0.05in}
\caption{Qualitative comparison of reconstructions of fixed samples of unconditional {{\method}} model (and its ablations) on CelebA.}
\label{fig:uncond_recon_celeba}
\vspace{0.15in}

\centering
\hspace*{\fill}
\begin{subfigure}{0.24\linewidth}
    \includegraphics[width=\linewidth]{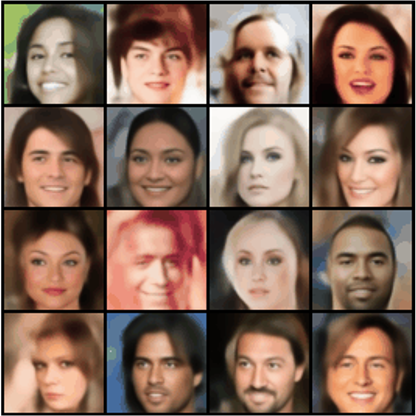}
    \caption{$L_{\tt recon}$ only (FID: 50.1)}
\end{subfigure}
\hspace*{\fill}
\begin{subfigure}{0.24\linewidth}
    \includegraphics[width=\linewidth]{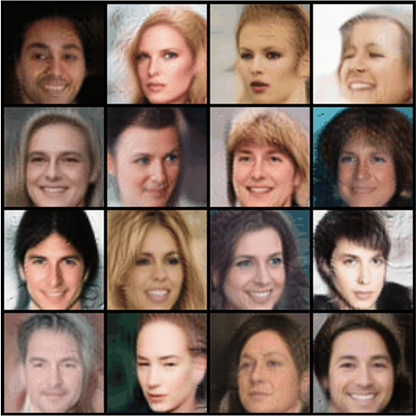}
    \caption{+ Adv. similarity (FID: 25.1)}
\end{subfigure}
\hspace*{\fill}
\begin{subfigure}{0.24\linewidth}
    \includegraphics[width=\linewidth]{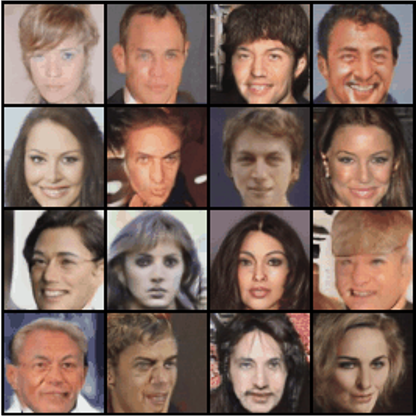}
    \caption{+ Projected (FID: 6.91)}
\end{subfigure}
\hspace*{\fill}
\vspace{-0.05in}
\caption{Qualitative comparison on uncurated random samples from unconditional {{\method}} model (and its ablations) on CelebA.}
\label{figure:uncond_sampling_celeba}
\vspace{0.15in}

\centering
\hspace*{\fill}
\begin{subfigure}{0.38\linewidth}
    \includegraphics[width=\linewidth]{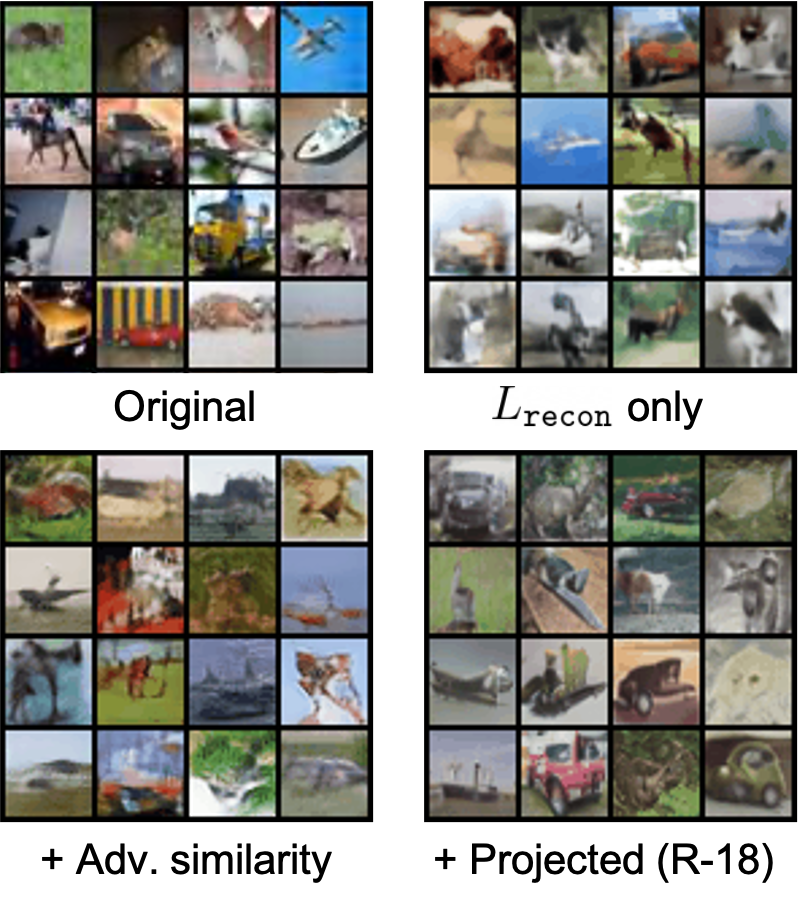}
    \caption{Reconstruction}
    \label{fig:original_cifar}
\end{subfigure}
\hspace*{\fill}
\begin{subfigure}{0.38\linewidth}
    \includegraphics[width=\linewidth]{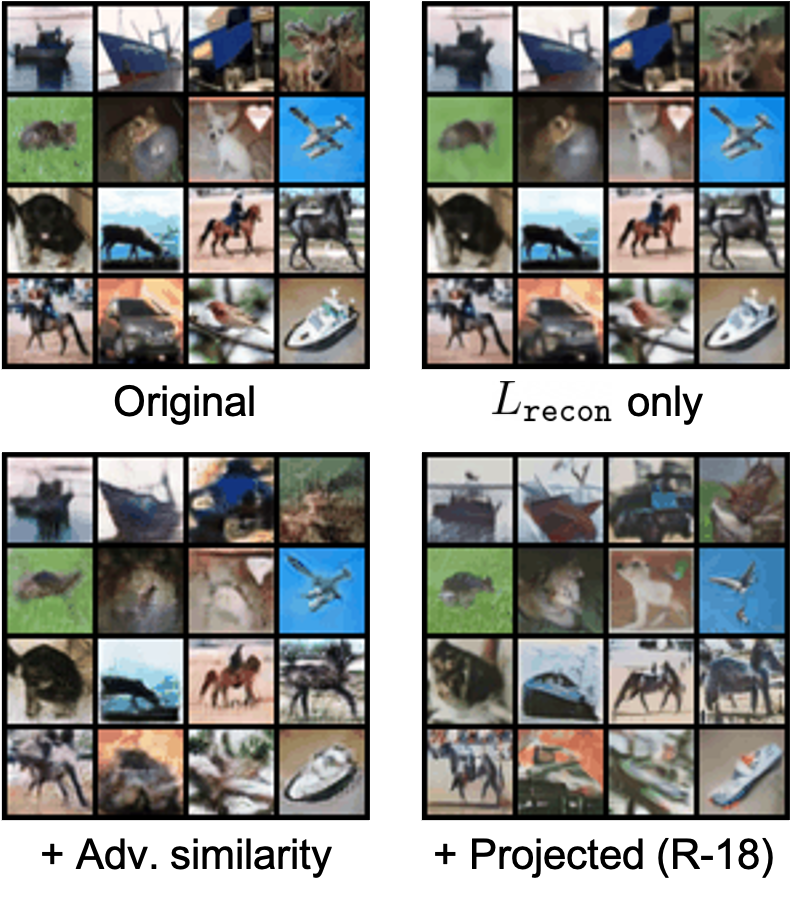}
    \caption{Sampling}
    \label{fig:recon_cifar}
\end{subfigure}
\hspace*{\fill}
\vspace{-0.05in}
\caption{Qualitative comparison on uncurated random samples from unconditional {{\method}} model (and its ablations) on CelebA.}
\label{figure:uncond_sampling_cifar}
\end{figure*}

\clearpage
\FloatBarrier
\section{Application to model debugging}

\begin{figure*}[ht]
  \centering
  \includegraphics[width=\linewidth]{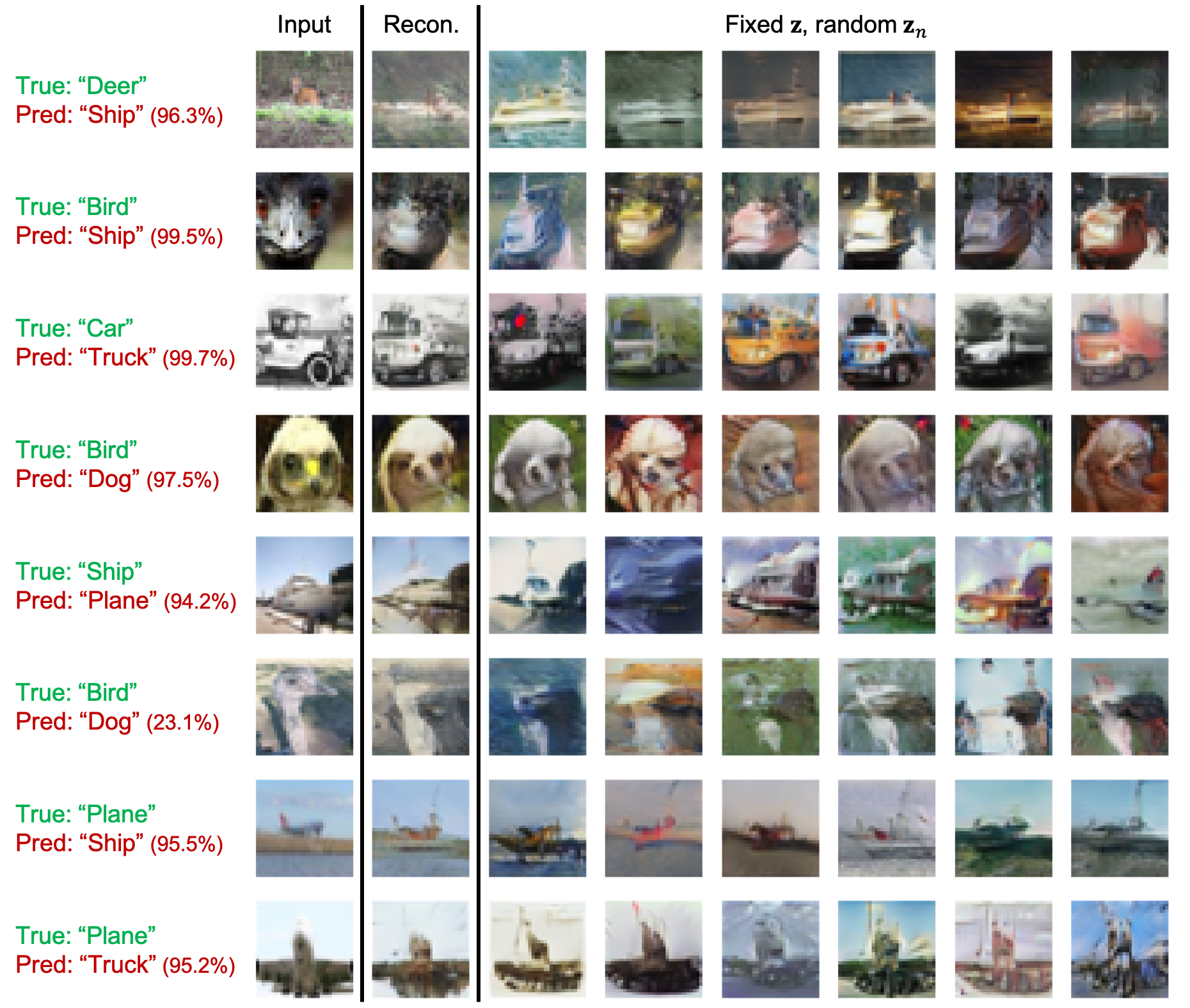}
  \caption{Qualitative comparison between (a) original input (the leftmost column), (b) its reconstruction (the second column), and (c) its further reconstructions with random nuisance $\rvz_n$ (the remaining columns), examined for test samples misclassified by a CIFAR-10 {{\method}} model.}
  \label{fig:model_debugging}
\end{figure*}

To further understand how the proposed {{\method}} model internally works with its representation $\rvz$ and $\rvz_n$, we examine an {{\method}} model trained on CIFAR-10 to analyze how the model reconstruct given inputs when the model incorrectly classifies them. Specifically, Figure~\ref{fig:model_debugging} illustrates a subset of CIFAR-10 test samples misclassified by an {{\method}} model by comparing the original input with its reconstructed samples from the model. Overall, we observe that such a qualitative comparison can provide a useful signal to interpret model errors: it effectively visualizes which visual cues of a given input negatively affected the decision making process of the given model, also visualizing the closest (misclassified) realizations that the model decodes for a given representation, \ie what the model actually perceived. For example, for the test input given at the first row of Figure~\ref{fig:model_debugging}, one can observe that the model essentially ``ignored'' the tiny part that represent the true semantic, \ie the ``deer'', and reconstructed the remaining part as a ``ship''.

\end{document}